\documentclass[]{bytedance_seed}

\usepackage{amsmath,amsfonts,bm}

\def\eqref#1{equation~\ref{#1}}

\def\1{\bm{1}}

\def\mA{{\bm{A}}}
\def\mB{{\bm{B}}}

\def\mH{{\bm{H}}}

\def\mW{{\bm{W}}}
\def\mX{{\bm{X}}}

\def\mLambda{{\bm{\Lambda}}}

\def\mdelta{{\bm{\delta}}}

\DeclareMathAlphabet{\mathsfit}{\encodingdefault}{\sfdefault}{m}{sl}
\SetMathAlphabet{\mathsfit}{bold}{\encodingdefault}{\sfdefault}{bx}{n}

\newcommand{\softmax}{\mathrm{softmax}}

\DeclareMathOperator*{\argmin}{arg\,min}

\usepackage{subcaption} 
\usepackage{hyperref}
\usepackage{url}
\usepackage{multirow}
\usepackage{booktabs}
\usepackage{threeparttable}
\usepackage{graphicx}
\usepackage{pifont}
\usepackage[table]{xcolor}
\usepackage{float}

\newtheorem{proposition}{Proposition}
\newtheorem{definition}{Definition}

\usepackage{algorithm}
\usepackage{algorithmic}

\usepackage[toc,page,header]{appendix}

\title{Train Large, Deploy Compact: Structured Compression for Compact Low-Rank Adaptation}

\author[1,2,*,]{Xin Yu}
\author[1]{Cong Xie}
\author[1,*]{Xunmei Liu}
\author[1]{Ziyu Zhao}
\author[1]{Tiantian Fan}
\author[2,\dagger]{Lingzhou Xue}
\author[1,\dagger]{Zhi Zhang}

\affiliation[1]{ByteDance Seed}
\affiliation[2]{The Pennsylvania State University}

\contribution[*]{Intern work at ByteDance Seed.}
\contribution[\dagger]{Corresponding authors. \texttt{zhangzhi.joshua@bytedance.com, lingzhou@psu.edu}}

\abstract{
Low-rank adaptation (LoRA) has become a widely used paradigm for efficiently adapting large language models, yet its empirical performance often lags behind full fine-tuning because of its low-rank constraint. A key open question is how to obtain expressive low-rank adapters from over-parameterized spaces. We propose \textit{PrunedLoRA}, a structured pruning framework that trains in an over-parameterized low-rank adapter space and compresses it into a compact final adapter. Unlike prior approaches that impose a fixed low-rank constraint throughout training, \textit{PrunedLoRA} dynamically prunes less important components during fine-tuning and prevents their reactivation, enabling flexible and adaptive rank allocation. By minimizing the pruning-induced error in the overall loss, we derive fine-grained pruning and recovery updates for a gradient-based pruning strategy and provide an interpretation of the resulting objective. To our knowledge, we provide one of the first theoretical analyses of the robustness of structured pruning and show, in a stylized setting, that under weight perturbations gradient-based pruning is more robust than activation-based pruning with respect to the overall loss. Empirically, \textit{PrunedLoRA} remains effective across both memory regimes: under practical memory constraints, it improves over standard LoRA and DoRA and remains competitive with adaptive-rank variants; when larger initialization ranks are affordable, it further narrows the gap to full fine-tuning. It also outperforms existing structured pruning methods across diverse sparsity levels.
}

\date{\today}

\begin{document}

\maketitle

\section{Introduction}

   Low-rank adaptation (LoRA) \citep{hu2022lora} and its variants \citep{zhang2023adalora,liu2024dora,hayou2024lora+} have emerged as a widely used paradigm for adapting large-scale foundation models \citep{sidahmed2024parameter, luo2023lcm, zhao2025each}. By injecting trainable low-rank matrices into the pre-trained model, LoRA enables efficient fine-tuning with minimal training overhead and no additional inference latency. Despite its efficiency, LoRA often lags behind full fine-tuning (FFT) in practical performance. Existing attempts to bridge this gap fall into two categories. The first line of work strictly follows LoRA's memory constraint, so exploring over the full parameter space is inadmissible~\citep{hayou2024lora+,yen2024lora,kalajdzievski2023rank,chen2025lora}. Learning within the low-rank space is always difficult to utilize the powerful representation of FFT \citep{zhang2025lora,hao2024flora}. The second line of work enables full-parameter learning \citep{zhao2024galore,hesubspace, liao2024galore} through projection techniques to compress and decompress gradients and weights. While these over-parameterized methods improve performance, they ultimately output fine-tuned full models rather than preserving a shared base model with lightweight, task-specific low-rank adapters. As a result, for the inference period, these approaches with full-parameter learning are less efficient, since each task requires storing a full model. In contrast, if we obtain low-rank adapters for different tasks, inference time and memory cost can be significantly reduced \citep{yang2025mtl, liao2025dynamic, feng2024mixture}. Therefore, the key question remains open: \emph{how can we find highly representative low-rank adapters from an over-parameterized setting for better performance?}

Empirically, we observe that increasing the rank of LoRA improves performance, in some cases approaching that of FFT (see Fig.~\ref{fig_rank_increase_lora} in Subsection \ref{sec_motivation}), a trend also reported in prior work~\citep{wang2024lorapro,hu2022lora}. This suggests that LoRA with a larger rank has sufficient representational capacity. Motivated by this observation, we consider initializing LoRA with a larger rank to ensure sufficient representational capacity, and then reducing the size of the model during fine-tuning to obtain a lightweight low-rank adapter. This strategy preserves the expressive power of an over-parameterized initialization while maintaining efficiency.

To realize this idea, we next turn to structured pruning \citep{lecun1989optimal,hassibi1993optimal,denil2013predicting, zhu2017prune}, a principled approach for reducing the model size by removing entire sub-components, such as rows or columns, from the model’s weight matrices. Two main categories of structured pruning have been widely studied: gradient-based methods \citep{molchanov2016pruning, yang2022gradient, ma2023llm} and activation-based methods \citep{frantar2023sparsegpt, kurtic2023ziplm, zhao2022adaptive}. Empirical evidence (e.g., \cite{nonnenmacher2021sosp}) suggests that gradient-based approaches focus more on global information and tend to be more stable with respect to the overall loss under weight perturbations. However, from a theoretical perspective, a clear comparison between these two classes of methods, particularly regarding how weight perturbations affect the overall loss, remains largely unexplored. To further mitigate the influence of pruning, \cite{frantar2023sparsegpt,kurtic2023ziplm,singh2020woodfisher} proposes updating weights after pruning, inspired by \emph{Optimal Brain Surgeon} \citep{hassibi1992second}. While these approaches investigate how to scale second-order methods to deep neural networks, they, as the original work~\cite{hassibi1992second}, leave open a deeper understanding of the pruning metric, known as the ``saliency" term in \emph{Optimal Brain Surgeon}.


In this work, with the goal of narrowing the gap between LoRA and full fine-tuning, we propose \textit{PrunedLoRA}, a structured pruning framework for learning compact low-rank adapters from an over-parameterized adapter space. Unlike existing methods that operate under a fixed low-rank budget throughout training, \textit{PrunedLoRA} begins training in a more expressive adapter space and progressively compresses it into a lightweight final adapter. A key advantage of \textit{PrunedLoRA} is that it remains effective across different memory regimes: under practical memory constraints, it yields stronger compact adapters than LoRA and its variants at the same deployment rank; when larger initialization ranks are affordable, it can further narrow the gap to full fine-tuning. The latter regime is particularly relevant when the objective is not merely to minimize one-time training cost, but to obtain the strongest possible compact adapter for deployment. In multi-tenant serving settings, the long-term inference footprint of many task-specific adapters can outweigh the training budget, which makes higher-quality compact adapters especially valuable \citep{punica,slora,lora_without_regret}. For the theoretical analysis of structured pruning, we consider a toy model of self-attention \citep{vaswani2017attention} and show that gradient-based pruning is more robust to weight perturbations in terms of overall loss than activation-based pruning in this stylized setting. We further discuss how the same intuition may extend to broader contexts. In addition, we provide a fine-grained analysis of pruning selection and weight update for weight matrices in a second-order gradient-based pruning strategy, which deepens the understanding of the pruning metric (the ``saliency" term in Eq. 5 of \cite{hassibi1992second}) in this class of second-order pruning methods.

We summarize our contribution as follows:

\begin{itemize}
    \item We propose \textit{PrunedLoRA}, a new framework that identifies highly representative low-rank adapters by structured pruning from an over-parameterized initialization, increasing representational capacity while retaining efficiency. Unlike prior approaches with a fixed low-rank budget, \textit{PrunedLoRA} enforces the low-rank constraint only at the end of fine-tuning, enabling flexible and adaptive rank allocation during training.
    \item We establish a theoretical analysis of the robustness of two major structured pruning approaches for large language models. Using a toy self-attention model, we show that gradient-based pruning is more robust to weight perturbations in terms of overall loss than activation-based pruning, and we further discuss how this intuition can extend to broader settings.

    \item We conduct extensive experiments across supervised fine-tuning tasks spanning mathematical reasoning, code generation, commonsense reasoning, and natural language understanding, showing that \textit{PrunedLoRA} can further narrow the gap between LoRA and FFT. Across different sparsity levels from $50\%$ to $93\%$ and across various pruning tasks (including both dynamic and one-shot pruning), our method consistently outperforms existing structured pruning methods.
\end{itemize}

\section{Methods}\label{sec_methods}

\subsection{Motivation}\label{sec_motivation}

\begin{figure}[htbp]
    \centering
    \begin{minipage}{0.46\linewidth}
        \centering
        \includegraphics[width=\linewidth]{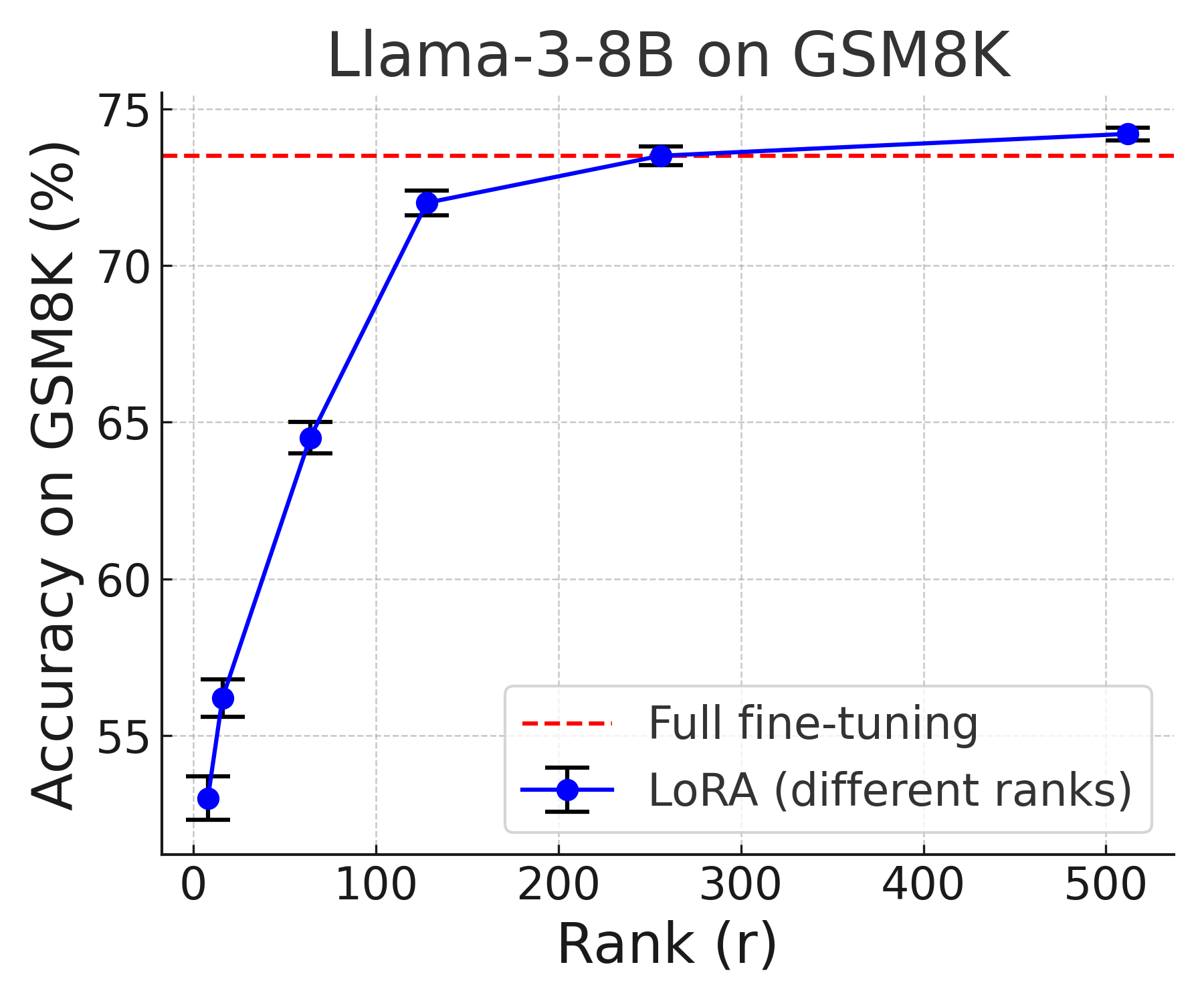}
    \end{minipage}
    \hspace{0.05\linewidth} 
    \begin{minipage}{0.46\linewidth}
        \centering
        \includegraphics[width=\linewidth]{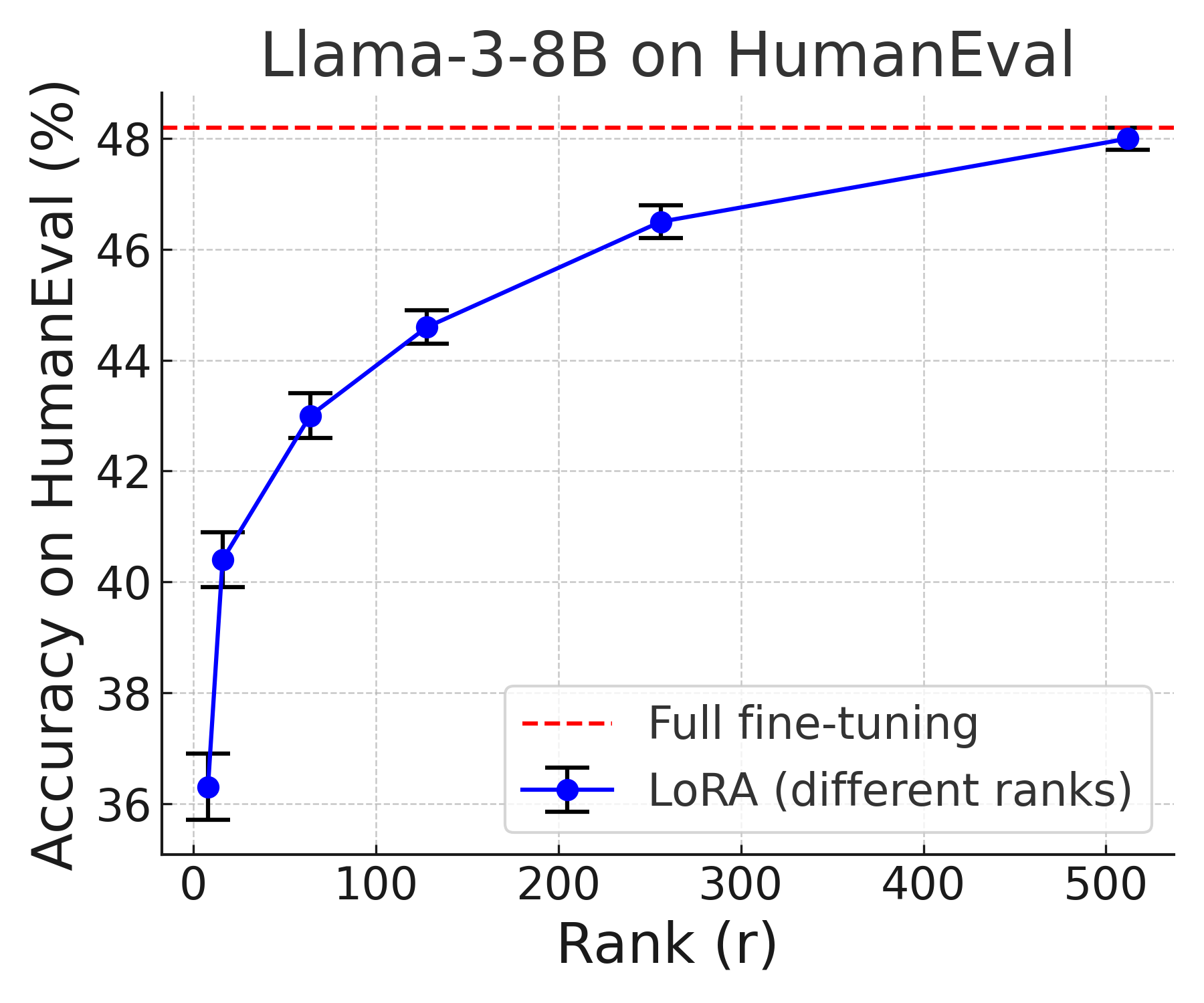}
    \end{minipage}
    \caption{Performance of standard LoRA \citep{hu2022lora} on GSM8K \citep{cobbe2021training} and HumanEval \citep{chen2021evaluating} with different ranks compared to full fine-tuning. Note that the method of full fine-tuning does not involve the initial rank, and we draw a red line here solely for comparison.}
    \label{fig_rank_increase_lora}
\end{figure}

\textbf{Motivation 1: Higher ranks lead to better performance.} As illustrated in Figure~\ref{fig_rank_increase_lora}, employing higher ranks in LoRA consistently leads to improved empirical performance on both GSM8K and HumanEval (see Sec.~\ref{sec_Experiment} for details). Notably, as the rank increases, the performance gradually converges toward that of full fine-tuning. This observation motivates our approach: rather than fixing LoRA to a small rank at the outset, we initialize with a sufficiently large rank—providing a number of trainable parameters close to full fine-tuning—and then progressively prune it to a smaller rank. Such a strategy may preserve most of the performance gains in over-parameterized settings while ultimately producing a memory-efficient low-rank adaptation.

PrunedLoRA is not designed solely to rely on very large initial ranks. A key motivation is that the paradigm is effective even under memory-constrained initializations that are typical in practice. Concretely, when the initial rank must remain moderate (e.g., $128$), training in a slightly over-parameterized adapter space and then pruning to a compact target rank yields consistently stronger adapters than training directly at the same target rank. For example, pruning from rank $64$ to $8$, or from rank $128$ to $64$, improves performance over standard LoRA trained at rank $8$ or $64$ with comparable deployment cost. This suggests that the benefit of PrunedLoRA comes from better searching and distilling the low-rank subspace during fine-tuning, rather than being tied to a single high-rank setting.

\textbf{Motivation 2: $\mA$ and $\mB$ in LoRA control the low-rank spaces.}
For the factor matrices $\mA \in \mathbb{R}^{r \times n}$ and $\mB \in \mathbb{R}^{m \times r}$, we observe that the columns of $\mB$ correspond to the column space of the original update $\Delta \mW$, while the rows of $\mA$ represent its row space \citep{yu2025altlora}. Therefore, they capture row-wise and column-wise correlations separately. As we discuss next, pruning these sub-modules instead of the full matrix reduces the computational cost and substantially simplifies second-order structured pruning.
This observation provides a natural bridge from the over-parameterized training strategy in Motivation 1 to our final compression procedure. If the goal is to reduce the effective rank while preserving the learned low-rank subspace, then pruning $\mA$ and $\mB$ directly is the natural structured operation, since it removes components of the adapter space itself rather than heuristically modifying the full update matrix.

\subsection{The Robustness of Gradient-Based Structured Pruning} \label{sec_pruning}

\textbf{Activation-Based \textit{vs.} Gradient-Based Structured Pruning.} Pruning induces perturbations to the weights across layers of large language models, which in turn modifies the overall loss and may deteriorate performance \citep{frantar2023sparsegpt,yang2023theoretical}. Within structured pruning, activation-based approaches optimize a reconstruction objective, while gradient-based approaches directly use the gradient of the overall loss to determine which columns to prune. Intuitively, gradient-based methods preserve global correlations more faithfully \citep{nonnenmacher2021sosp}, suggesting greater robustness to weight perturbations. We formalize this comparison below; detailed background and derivations are deferred to the supplementary material.

\begin{proposition}[Unofficial Statement] \label{proposition_comp_unofficial}
    Suppose that, under activation-based and gradient-based pruning strategies, each module in a single attention module satisfies a given perturbation error. The error in the loss function would be linear w.r.t. perturbation error under different pruning strategies, but the error of activation-based methods depends on the magnitude of each module.
\end{proposition}

Proposition~\ref{proposition_comp_unofficial} suggests that activation-based methods can introduce higher sensitivity in the overall loss. This is consistent with the insight that activation-based methods do not directly capture how weight changes affect global correlation \citep{das2023beyond}. In the supplementary material, we also provide experimental support for Proposition \ref{proposition_comp_unofficial} using a self-attention model trained on a linear regression task. Our robustness analysis is derived under stylized assumptions and approximations to make the comparison tractable, and extending analogous guarantees to realistic large-scale training dynamics remains an open direction. We begin by formulating the gradient-based pruning problem for the columns of a full weight matrix, which can be interpreted as pruning the columns of matrix $\mB$ (or, symmetrically, the rows of matrix $\mA$) alone.

\textbf{Problem formulation.} Our approach starts from the idea of applying a structured compression layer-wise, in a way that allows the layers to preserve most of their output characteristics. This setup is popular in the post-training quantization and unstructured pruning literature~\citep{frantar2023sparsegpt,tang2025darwinlm,wu2024iterative}, and can be implemented as follows. During fine-tuning, the gradient is non-trivial because it helps the fine-tuned model align with the downstream task data. Therefore, our setup is different from the literature in gradient-based pruning \citep{singh2020woodfisher, kurtic2022optimal}. We consider the perturbation of a single weight matrix $\mW \in \mathbb{R}^{m \times n}$ in a large language model. The pruned matrix is denoted as $\mW + \mdelta$, where the perturbation $\mdelta \in \mathbb{R}^{m \times n}$ corresponds to pruning the same weight indices across all rows, i.e., entire columns are removed. The update $\mdelta \in \mathbb{R}^{m \times n}$ is subject to the constraint that
\begin{align}\label{assumption_structure}
\mdelta_{:, \mathcal{M}_s} = - \mW_{:, \mathcal{M}_s}.
\end{align}
Here, $\mathcal{M}_s$ denotes the pruning mask that specifies the pruned column indices with sparsity s. Expanding the overall loss of the pruned model with weight matrix $\mW + \mdelta$ around $\mW$ yields
\begin{align}\label{eq:taylor_expansion}
\mathcal{L}(\mW + \mdelta) \approx\; & \mathcal{L}(\mW) +
\left\langle \nabla_\mW \mathcal{L}(\mW), \mdelta \right\rangle \\
 & + \frac{1}{2} tr(\textit{vec}(\mdelta)^\top \mathbf{H}\,\textit{vec}(\mdelta)),
\end{align}
which corresponds to the matrix-form second-order Taylor expansion, where $\textit{vec}(\mdelta)$ denotes the vectorization of the perturbation matrix. Noticeably, the Hessian matrix is $\mH\in \mathbb{R}^{mn\times mn}$, so the memory cost and the computational cost are extremely huge. To address the challenge, many existing methods propose to impose structural assumptions for the Hessian matrix $\mH$, such as diagonal or block-diagonal approximation~\citep{zhang2017block,hassibi1992second} and empirical Fisher~\citep{cho2015hessian,singh2020woodfisher}. With the goal of selecting columns in (\ref{assumption_structure}), it is critical to preserve the correlation among the columns of the weight matrix. Thus, with the standard assumption of row independence in \cite{kurtic2022optimal,frantar2023sparsegpt}, as a common technique for approximating the Hessian using gradients, we can approximate (\ref{eq:taylor_expansion}) by
\begin{align}\label{eq2:taylor_expansion}
\mathcal{L}(\mW + \mdelta) \approx\; & \mathcal{L}(\mW) +
\left\langle \nabla_{\mW} \mathcal{L}(\mW), \mdelta \right\rangle \\
& + \frac{1}{2} tr(\mdelta^\top \widehat{\mH}\mdelta),
\end{align}
where $\widehat{\mH} = (\nabla_{\mW} \mathcal{L}(\mW))^T \nabla_{\mW} \mathcal{L}(\mW) \in \mathbb{R}^{n\times n}$. Then, combining the pruned structure (\ref{assumption_structure}) with the analysis of perturbation in $\mW$ leads to the optimal pruning selection and weight update by solving the following problem:
\begin{align}\label{grad_objective_updateweight}
\mathcal{M}_s, \mdelta = \textit{argmin}_{\mathcal{M}_s,\mdelta} \; &
\left\langle \nabla_{\mW} \mathcal{L}, \mdelta \right\rangle +
\frac{1}{2} tr(\mdelta \widehat{\mH}\,\mdelta^\top ) \\
\textit{s.t.} \quad & \mdelta_{:, \mathcal{M}_s} = - \mW_{:, \mathcal{M}_s}.
\end{align}
Here, for simplicity, we denote $\nabla_{\mW}\mathcal{L}(\mW)$ as $\nabla_{\mW}\mathcal{L}$.
The optimal solution of $\mdelta$ in ( \ref{grad_objective_updateweight}) is 
\begin{align}\label{obj_update}
\mdelta
&= - \nabla_{\mW} \mathcal{L} \: \widehat{\mH}^{-1}
- \mW_{:, \mathcal{M}_s}
\left( (\widehat{\mH}^{-1})_{\mathcal{M}_s, \mathcal{M}_s} \right)^{-1}
(\widehat{\mH}^{-1})_{\mathcal{M}_s, :} \nonumber \\
&\quad + (\nabla_{\mW} \mathcal{L}  \widehat{\mH}^{-1})_{:, \mathcal{M}_s}
\left( (\widehat{\mH}^{-1})_{\mathcal{M}_s, \mathcal{M}_s} \right)^{-1}
(\widehat{\mH}^{-1})_{\mathcal{M}_s, :}.
\end{align}

\textbf{Interpretation for Algorithm Design.} Let us further analyze the update $\mdelta$ in (\ref{obj_update}). The first term in $\mdelta$ is a second-order Newton step. If there is no sparse masking, it would be the optimal update utilizing second-order momentum. As $P_{\mathcal{M}_s}\mdelta$ will only leave the second term in (\ref{obj_update}), which is a projection correction to ensure the pruned weights remain zero. Interestingly, it is dependent on the current weight $\mW$ and the mask $\mathcal{M}_s$ but independent of the gradient $\nabla_{\mW}\mathcal{L}$. The third term in (\ref{obj_update}) provides a dual variable compensation that projects the unconstrained Newton step into the feasible region. Once we get the closed-form solution of $\mdelta$, the corresponding pruning objective becomes
\begin{align}\label{obj_search_mask}
\min_{\mathcal M_s}\operatorname{tr}\!\Big(
&(\mathbf W-\nabla_{\mathbf W}\mathcal L\,\mathbf H^{-1})_{:\!,\mathcal M_s}
\big((\mathbf H^{-1})_{\mathcal M_s,\mathcal M_s}\big)^{-1} \nonumber\\
&(\mathbf W-\nabla_{\mathbf W}\mathcal L\,\mathbf H^{-1})_{:\!,\mathcal M_s}^{\top}
\Big).
\end{align}
Here, the pruning problem in (\ref{obj_search_mask}) is closely related to the ``saliency" term in \citep{hassibi1992second}. With the analysis of matrix weight, we provide \emph{an explicit interpretation} for second-order pruning strategies: \emph{we select the pruning mask that removes the columns whose post-update (second-order Newton update) values are least important under the Hessian-weighted quadratic metric}. Existing methods deriving from Optimal Brain Surgeon cannot provide a grounded interpretation from the ``saliency" term, as most of them focus on activation-reconstruction objectives~\citep{frantar2023sparsegpt,kurtic2023ziplm} or only analyze the one-dimensional weight vectors \citep{das2023beyond,singh2020woodfisher,kurtic2022optimal}. Therefore, our analysis enriches the understanding of the class of second-order pruning methods. 

\begin{figure}[t]
    \centering
    \includegraphics[width=1.0\linewidth]{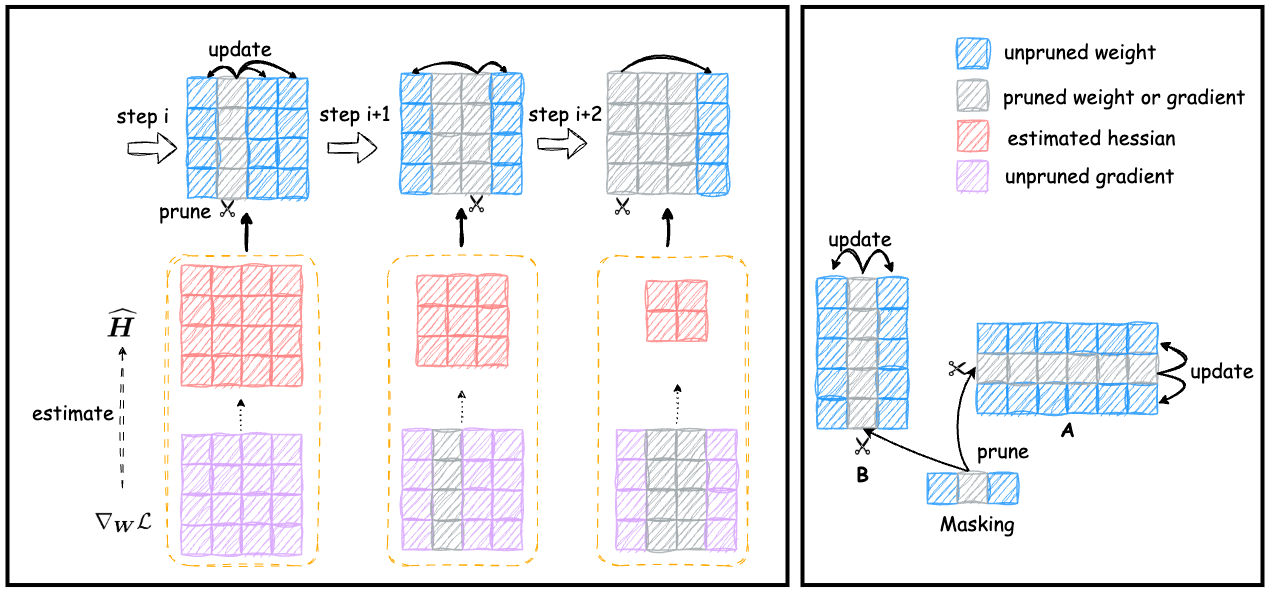}  
    \caption{
    \textbf{Left:} schematic of the dynamic pruning process, where the gradient and estimated Hessian determine the pruned columns and weight updates. 
    \textbf{Right:} design of \textit{PrunedLoRA}, where both adapter matrices $\mA$ and $\mB$ are jointly pruned under a masking scheme.}
    \label{fig:prunedlora_gsm8k}
\end{figure}

We summarize our solution in the supplementary material and present a schematic illustration of the workflow in the left of Figure \ref{fig:prunedlora_gsm8k}. In each pruning step, the pruning indices are determined by the gradient and the estimated Hessian.

\subsection{PrunedLoRA} \label{sec_pruning_lora}
\begin{algorithm}[ht]
\caption{\textbf{PrunedLoRA}: structured pruning for Low-rank Adapters from over-parameterized spaces. 
We prune LoRA matrices $(A,B)$ with column sparsity $s$ on $B$ (and corresponding row sparsity s on $A$) 
given gradients $(\nabla_A \mathcal{L}, \nabla_B \mathcal{L})$ and Hessian estimates $(\widehat{H}_A,\widehat{H}_B)$.}
\label{alg:prunedlora}
\begin{algorithmic}[1]

\STATE \textbf{Step 1: Search pruning mask.} 
\begin{align}
\begin{split}
\arg\min_{\mathcal{M}_s} \;
& \mathrm{tr}\!\left(
\widetilde{B}_{:,\mathcal{M}_s}
\big((\widehat{H}_B^{-1})_{\mathcal{M}_s,\mathcal{M}_s}\big)^{-1}
\widetilde{B}_{:,\mathcal{M}_s}^\top
\right) 
+ \\
&
\mathrm{tr}\!\left(
\widetilde{A}_{\mathcal{M}_s,:}^\top
\big((\widehat{H}_A^{-1})_{\mathcal{M}_s,\mathcal{M}_s}\big)^{-1}
\widetilde{A}_{\mathcal{M}_s,:}
\right),
\end{split}
\end{align}
where $\widetilde{A} = A - \widehat{H}_A^{-1}\nabla_A \mathcal{L}, \;
\widetilde{B} = B - \nabla_B \mathcal{L}\,\widehat{H}_B^{-1}$.

\STATE \textbf{Step 2: Compute optimal updates.} 
\STATE Given $\mathcal{M}_s$, compute
\begin{align*}
\delta_B ={}& -\nabla_B \mathcal{L}\,\widehat{H}_B^{-1}
- \widetilde{B}_{:,\mathcal{M}_s}
\big((\widehat{H}_B^{-1})_{\mathcal{M}_s,\mathcal{M}_s}\big)^{-1} \\
&(\widehat{H}_B^{-1})_{\mathcal{M}_s,:},
\end{align*}
\begin{align*}
\delta_A ={}& -\widehat{H}_A^{-1}\nabla_A \mathcal{L}
- (\widehat{H}_A^{-1})_{:,\mathcal{M}_s}
\big((\widehat{H}_A^{-1})_{\mathcal{M}_s,\mathcal{M}_s}\big)^{-1} \\
&\widetilde{A}_{\mathcal{M}_s,:}.
\end{align*}
\STATE Set $A \gets A + \delta_A$, \; $B \gets B + \delta_B$.

\STATE \textbf{Step 3: Update LoRA adapters with standard optimizer in fine-tuning.} 

\STATE \textbf{Step 4: Iterate or finalize.} 
\STATE If multi-round pruning is desired, repeat Steps 1--3 until the target rank is reached. Otherwise, output $(A,B)$.

\end{algorithmic}
\end{algorithm}

In this part, we propose our structured pruning strategy, termed \textit{PrunedLoRA}. Inspired by \emph{Motivation 1}, we dynamically prune adapters $\mA$ and $\mB$ from high-parameter spaces.

Different from prior work such as AdaLoRA~\citep{zhang2023adalora}, which enforces an average rank budget and dynamically selects ranks from a small predefined set, our method allows training to start in a more expressive adapter space and only imposes the final compact rank after pruning. Besides, structurally pruning the columns and rows of a full weight matrix causes high computational overhead, as we highlight in Eq. (\ref{eq:taylor_expansion}). However, with \emph{Motivation 2}, we can efficiently detect the row-wise and column-wise correlation by pruning the low-rank spaces of $\mA$ and $\mB$ together. With the goal of reducing the rank of the matrix, structured pruning of the decomposed sub-modules is more efficient.

With the standard argument in Sec~\ref{sec_pruning}, the pruning problem for low-rank adaptation $\mA$ and $\mB$ is



\begin{align}
    \begin{split}
         \textit{argmin}_{\mathcal{M}_s,\mdelta_{\mA} , \mdelta_{\mB}} \; & \left\langle \nabla_{\mA} \mathcal{L}, \mdelta_{\mA} \right\rangle + \frac{1}{2} tr(\mdelta_{\mA}^\top \, \widehat{\mH}_{\mA}\,\mdelta_{\mA}) \\
         & + \left\langle \nabla_{\mB} \mathcal{L}, \mdelta_{\mB} \right\rangle + \frac{1}{2} tr(\mdelta_{\mB} \widehat{\mH}_{\mB}\,\mdelta_{\mB}^\top) \\
         & \textit{s.t.} \quad (\mdelta_{\mB})_{:, \mathcal{M}_s} =  -{\mB}_{:, \mathcal{M}_s}, \\
         & \phantom{\textit{s.t.} \quad} (\mdelta_{\mA})_{\mathcal{M}_s,:}= -{\mA}_{\mathcal{M}_s,:}.
    \end{split}
\end{align}

Here, the mask $\mathcal{M}_s$ simultaneously controls the column sparsity of $\mB$ and the row sparsity of $\mA$. Consequently, the Hessian estimates $\widehat{\mH}_\mA$ and $\widehat{\mH}_\mB$ are computed with different purposes: to capture the column-wise correlations of $\mB$ and the row-wise correlations of $\mA$, respectively. Following the standard derivation in Sec ~\ref{sec_pruning}, our pruning strategy for reducing high-rank matrices $\mA$ and $\mB$ to a low-rank adaptation begins by determining the optimal pruning mask via

\begin{align}\label{obj_search_mask_lora}
\argmin_{\mathcal{M}_s} \; &
\operatorname{tr} \left(\widetilde{\mB}_{:, \mathcal{M}_s}
((\widehat{\mH}_{\mB})^{-1}_{\mathcal{M}_s, \mathcal{M}_s})^{-1}
\widetilde{\mB}_{:, \mathcal{M}_s}^T\right) \nonumber \\
&+ \operatorname{tr}\left(\widetilde{\mA}^T_{\mathcal{M}_s,:}
((\widehat{\mH}_{\mA})^{-1}_{\mathcal{M}_s, \mathcal{M}_s})^{-1}
\widetilde{\mA}_{\mathcal{M}_s,:}\right).
\end{align}
where $\widetilde{\mA}  = \mA - (\widehat{\mH}_{\mA})^{-1}\nabla_{\mA}\mathcal{L}, \:  \widetilde{\mB}  = \mB -  \nabla_{\mB}\mathcal{L}\:(\widehat{\mH}_{\mB})^{-1}$. After selecting the pruning indices, we update $\mA$ and $\mB$ as (\ref{obj_update_AB}) to minimize the perturbation error in the loss.
\begin{align}\label{obj_update_AB}
\begin{split}
\mdelta_{\mB} 
 & = - \nabla_{\mB} \mathcal{L} \: \widehat{\mH}_{\mB}^{-1}
 - \widetilde{\mB}_{:,\mathcal{M}_s}
\big((\widehat{\mH}_{\mB}^{-1})_{\mathcal{M}_s, \mathcal{M}_s}\big)^{-1}
(\widehat{\mH}_{\mB}^{-1})_{\mathcal{M}_s, :} \\
\mdelta_{\mA} & = -  \widehat{\mH}_{\mA}^{-1} \: \nabla_{\mA} \mathcal{L}
-  (\widehat{\mH}_{\mA}^{-1})_{:,\mathcal{M}_s}
\big((\widehat{\mH}_{\mA}^{-1})_{\mathcal{M}_s,\mathcal{M}_s}\big)^{-1}
\widetilde{\mA}_{\mathcal{M}_s,:}.
\end{split}
\end{align}

\textbf{Complexity.} For \textit{PrunedLoRA}, the pruning procedure begins with an initial rank much smaller than $\min\{m,n\}$ and progressively reduces the rank until reaching the target level. Since pruning is performed only for a limited number of steps, the additional cost introduced by the pruning operations remains moderate. Moreover, even for the initialization of LoRA with high rank, the computational complexity of pruning is $\mathcal{O}(r^3)$ and much smaller than that of matrix multiplication, i.e., $\mathcal{O}(\max\{m^2r, n^2r\})$. The computational time of pruning is mild. For instance, we isolate and measure the wall-clock time required solely for the
pruning procedure with initial rank $r = 512$ on the GSM8K datasets using the pretrained Llama-3-8B base model. The time consumed by pruning is 13 min (6.40$\%$ of the overall training time), demonstrating that the additional cost introduced by our
second-order pruning strategy is modest in practice. Consequently, our method maintains a computational cost comparable to that of existing low-rank adaptation approaches \citep{yen2024lora,yu2025altlora}.

\section{Experiment}
\label{sec_Experiment}
In this section, we evaluate \textit{PrunedLoRA} across supervised fine-tuning tasks in dialogue generation, mathematical reasoning, code generation, and natural language understanding. We focus on whether training in an over-parameterized adapter space and pruning during fine-tuning can consistently improve the quality of the final compact adapter. Additional task details, commonsense reasoning results, and ablations are provided in the supplementary material.

\textbf{Baselines.} We compare \textit{PrunedLoRA} with full fine-tuning, vanilla LoRA~\citep{hu2022lora}, two widely used LoRA variants (DoRA~\citep{liu2024dora} and AdaLoRA~\citep{zhang2023adalora}), and structured pruning baselines including SparseGPT and LLM-Pruner.

In addition to fine-tuning baselines, we also compare against existing structured pruning approaches for low-rank adaptation. Gradient-based pruning includes our method, which jointly optimizes parameter updates and pruning structure, as well as the widely used importance-score pruning strategy employed in LLM-Pruner~\citep{ma2023llm}. Activation-based pruning determines the pruning structure based on input activation statistics, as exemplified by ZipLM~\citep{kurtic2023ziplm} and SparseGPT~\citep{frantar2023sparsegpt}.

\subsection{Experiments on Supervised Fine-tuning}\label{sec_sup}

\textbf{Implementation Details.} 
To ensure a fair comparison, we align our experimental setup with the literature \citep{wang2024loraga, wang2024lorapro,yu2025altlora}. We fine-tune the model using the AdamW optimizer with hyperparameters $\beta_1 = 0.9, \beta_2 = 0.999$, and weight decay set to 0. We implement a cosine learning rate schedule with a warm-up ratio of 0.03. LoRA is applied to all linear modules, excluding the embedding layer and normalization layer. All experiments are conducted on NVIDIA H100 GPUs. To obtain a reliable estimate of model performance, we perform three runs with different random seeds and report the average and standard deviation of the results.

For dialogue generation, mathematical reasoning, and code generation tasks, we set the target LoRA rank $r \in \{8, 64\}$ and search the scaling factor $\alpha$ over $\{r/2, r, 2r\}$. For structural pruning methods, the initial ranks are set to 64 and 128 as default, respectively, to ensure a sufficiently expressive pre-pruning space. For the learning rate, we perform a grid search over $\{2\times10^{-4}, 5\times10^{-5}, 2\times10^{-5}\}$. This hyperparameter range fully covers the settings used in prior work~\citep{wang2024loraga, wang2024lorapro,yu2025altlora}. We use a sequence length of 1024 and a macro batch size of~32.

For natural language understanding tasks, we fine-tune the T5-base model \citep{raffel2020exploring} with learning rate of $1e{-4}$ and target LoRA rank $8$, using a sequence length of 128 and a batch size of 32. For structural pruning methods, we set the initial LoRA rank to 64. For DoRA, we adopt a learning rate of $2\times10^{-4}$, while for AdaLoRA, we follow prior work and use $5\times10^{-4}$. We keep these task-specific settings consistent with the reference implementations in \cite{wang2024loraga,wang2024lorapro,yu2025altlora} so that each baseline is evaluated under its standard tuning regime.

\textbf{Results on Natural Language Generation.} Following the configuration used in \citep{wang2024loraga, wang2024lorapro,yu2025altlora}, we evaluate the performance of PrunedLoRA on large language models, focusing on dialogue generation, mathematical reasoning and code generation capabilities. Detailed task descriptions are provided in the supplementary material.

\begin{table*}[h]
\centering
\small
\begin{threeparttable}
\begin{tabular}{l c c c c}
\toprule
\textbf{Method} & \textbf{Target Rank} & \textbf{MT-Bench} $\uparrow$ & \textbf{GSM8K} $\uparrow$ & \textbf{HumanEval} $\uparrow$ \\
\midrule
PreTrain & -- & 5.89 $\pm$ 0.04 & 51.34 $\pm$ 1.38 & 34.21 $\pm$ 0.23 \\
Full FT & -- & \textbf{6.31 $\pm$ 0.03} & \underline{73.48 $\pm$ 0.42} & \underline{48.28 $\pm$ 0.03} \\
\midrule
\multirow{2}{*}{LoRA}    
    & 8  & 6.01 $\pm$ 0.05 & 65.27 $\pm$ 0.13 & 39.23 $\pm$ 0.78 \\
    & 64 & 6.19 $\pm$ 0.03 & 69.21 $\pm$ 0.36 & 42.88 $\pm$ 0.34 \\
\midrule
\multirow{2}{*}{DoRA}    
    & 8  & 6.07 $\pm$ 0.02 & 67.08 $\pm$ 0.31 & 41.28 $\pm$ 0.39 \\
    & 64 & 6.23 $\pm$ 0.03 & 70.43 $\pm$ 0.21 & 43.32 $\pm$ 0.29 \\
\multirow{2}{*}{AdaLoRA} 
    & 8  & 6.08 $\pm$ 0.03 & 71.24 $\pm$ 1.32 & 41.88 $\pm$ 1.15 \\
    & 64 & 6.12 $\pm$ 0.08 & 71.45 $\pm$ 1.37 & 42.34 $\pm$ 1.41 \\
\midrule
\multirow{2}{*}{SparseGPT} 
    & 8  (init r=64) & 6.09 $\pm$ 0.02 & 67.28 $\pm$ 0.29 & 41.43 $\pm$ 0.28 \\
    & 64 (init r=128)  & 6.16 $\pm$ 0.02 & 69.71 $\pm$ 0.48 & 43.82 $\pm$ 0.39 \\
\multirow{2}{*}{LLM-Pruner} 
    & 8  (init r=64) & 6.09 $\pm$ 0.03 & 69.88 $\pm$ 0.35 & 42.25 $\pm$ 0.32 \\
    & 64  (init r=128) & 6.18 $\pm$ 0.03 & 70.88 $\pm$ 0.45 & 44.38 $\pm$ 0.12 \\
\midrule
\multirow{4}{*}{PrunedLoRA} 
    & 8  (init r=64)                  & 6.14 $\pm$ 0.06 & 69.02 $\pm$ 0.53 & 42.32 $\pm$ 0.33 \\
    & 64  (init r=128)             & 6.19 $\pm$ 0.04 & 71.16 $\pm$ 0.24 & 44.32 $\pm$ 0.11 \\
    & 64 (init r=256)  & 6.23 $\pm$ 0.03 & 72.21 $\pm$ 0.45 & 46.21 $\pm$ 0.26 \\
    & 64 (init r=512)  & \underline{6.25 $\pm$ 0.06} & \textbf{74.88 $\pm$ 0.42} & \textbf{48.31 $\pm$ 0.24} \\
\bottomrule
\end{tabular}
\end{threeparttable}
\caption{Performance comparison of fine-tuning and pruning baselines on MT-bench, GSM8K and HumanEval benchmarks for Llama-3-8B-Base Model. \textbf{Bold} indicates the best result, \underline{underline} represents the second-best one. ($\uparrow$:\; higher values indicate better performance)}
\label{tab_sft_score_3_8B}
\end{table*}

Table \ref{tab_sft_score_3_8B} highlights that \textit{PrunedLoRA} is effective in both training-memory-constrained and less constrained regimes. Under moderate initialization budgets, it already improves the quality of compact adapters at the same deployed rank. For target rank 8, \textit{PrunedLoRA} with init $r=64$ achieves 69.02 on GSM8K and 42.32 on HumanEval, substantially outperforming LoRA at rank 8 (65.27 and 39.23) and DoRA at rank 8 (67.08 and 41.28), while remaining close to AdaLoRA on GSM8K (71.24) and exceeding it on HumanEval (42.32 vs.\ 41.88). Similarly, for target rank 64, \textit{PrunedLoRA} with init $r=128$ reaches 71.16 on GSM8K and 44.32 on HumanEval, improving over LoRA (69.21 and 42.88) and DoRA (70.43 and 43.32), and staying competitive with AdaLoRA (71.45 and 42.34). These results show that the train-then-compress strategy is useful even when the initial rank is only moderately larger than the final deployment rank.

When larger training budgets are affordable, scaling up the initialization rank further narrows the gap to full fine-tuning. With target rank 64, \textit{PrunedLoRA} initialized at 256 reaches 72.21 on GSM8K and 46.21 on HumanEval, and initialization at 512 further improves the scores to 74.88 and 48.31. These results bring \textit{PrunedLoRA} very close to full fine-tuning on GSM8K and HumanEval, where full fine-tuning attains 73.48 and 48.28, respectively. We observe the same pattern beyond the main generation tasks: detailed commonsense reasoning results are provided in the supplementary material. This shows that \textit{PrunedLoRA} can use a highly expressive training-time adapter space to approach full fine-tuning performance while still producing a compact final adapter.

\begin{table*}[t]
\centering
\small
\begin{threeparttable}
\setlength{\tabcolsep}{6pt}
\begin{tabular}{l|c|cc|c|c}
\toprule
Method &  Rank & \textbf{Before (\%)} & \textbf{After (\%)} & Memory & Time \\
\midrule
Full FT   & full rank  & 100.00 & 100.00 & $\sim$ 8 $\times$ 40G & 4h 23min \\ 
LoRA      & 8  & 0.11 & 0.11  & $\sim$ 8 $\times$ 17G  & 2h 27min \\
LoRA      & 64  & 0.84 & 0.84 & $\sim$ 8 $\times$ 18G  & 2h 28min \\
DoRA      & 64  & 0.89 & 0.89 & $\sim$ 8 $\times$ 18G  & 2h 34min \\
AdaLoRA   & 64  & 0.84 & 0.84 & $\sim$ 8 $\times$ 19G  & 2h 41min \\
\midrule
PrunedLoRA (init r = 64)  & 8  & 0.84 & 0.11 & $\sim$ 8 $\times$ 19G & 2h 29min \\
PrunedLoRA (init r = 128) & 64 & 1.68 & 0.84 & $\sim$ 8 $\times$ 20G & 2h 31min \\
PrunedLoRA (init r = 256) & 64 & 3.36 & 0.84 & $\sim$ 8 $\times$ 22G & 2h 47min \\
PrunedLoRA (init r = 512) & 64 & 6.71 & 0.84 & $\sim$ 8 $\times$ 28G & 3h 23min \\
\bottomrule
\end{tabular}
\end{threeparttable}
\caption{Comparison of trainable parameter ratios, peak memory cost (per GPU on 8$\times$H100 with FSDP), and training time across different fine-tuning methods.}
\label{tab:lora_pro_comparison}
\end{table*}

\textbf{Memory and Time Costs.}  In Table \ref{tab:lora_pro_comparison}, we compare the percentage of trainable parameters (before and after pruning), peak memory cost and training time of our methods with full fine-tuning, LoRA, DoRA, and AdaLoRA on the math task and Llama-3-8B model. We measure memory cost using 8 H100 GPUs with batch size 1 following \cite{wang2024lorapro}. As the number of pruning steps is quite small relative to the overall fine-tuning process, the training time remains comparable. For example, for an initial rank of 512 in \textit{PrunedLoRA}, the total time spent on pruning is 23 minutes. From Table~\ref{tab:lora_pro_comparison}, we make two key observations: (1) Even with a very high initial LoRA rank—such as 512—the peak memory consumption of \textit{PrunedLoRA} remains substantially lower than that of full fine-tuning. (2) The additional computation introduced by the pruning procedure incurs only a mild overhead beyond the standard LoRA forward and backward passes. Empirically, LoRA with rank 64 requires 2h 28 min, while PrunedLoRA (target rank 8, initial rank 64) completes in 2h 29 min under the same setup. This demonstrates that the pruning step adds only a mild runtime cost. At the same time, PrunedLoRA trades additional training-time memory/compute for better low-rank adapters: very high-rank initialization may be infeasible under strict training-memory budgets, and lower-rank initializations may correspondingly yield smaller gains. Moreover, the deployment-side amortization argument is strongest in settings that reuse many adapters over time; in single-task deployments, this advantage is weaker.

\textbf{Results on Natural Language Understanding.} 
We observe the same two-regime pattern on GLUE: under moderate training budgets, \textit{PrunedLoRA} improves over LoRA-style baselines, and under larger training budgets it further narrows the gap to full fine-tuning. Detailed per-task GLUE results are provided in the supplementary material.

\subsection{Experiments on Ablation Study}\label{Sec_abl}

We defer detailed ablations to the supplementary material. There we study initialization rank, scaling factor, aggressive target-rank compression, and pruning schedules. Across these settings, the conclusions are consistent with the main text: \textit{PrunedLoRA} remains robust under stronger compression and preserves most of the benefit of training in a larger adapter space.

\section{Conclusion}
In this work, we introduced \textit{PrunedLoRA}, a gradient-based structured pruning framework for obtaining efficient low-rank adapters from over-parameterized spaces. By formulating pruning as an optimization problem that explicitly minimizes the loss induced by weight perturbations, our method provides a theoretically grounded strategy for structured adapter compression. Comprehensive experiments on mathematical reasoning, code generation, and natural language understanding demonstrate that \textit{PrunedLoRA} consistently narrows the gap to full fine-tuning while retaining inference efficiency. Furthermore, across diverse sparsity levels, it achieves superior performance over existing structured pruning baselines, underscoring both its robustness and practical effectiveness.


\newpage

\bibliographystyle{plainnat}
\bibliography{paper}

\clearpage

\beginappendix

\section{The Use of LLMs}
LLMs were used to improve writing clarity and assist with code development. Specifically, LLMs assisted in improving the clarity, fluency, and grammatical correctness of the manuscript, including rephrasing sentences and ensuring consistent terminology. Additionally, LLMs helped generate auxiliary code and scripts for data processing, experimental setup, and result visualization. However, the core research ideas, technical contributions, experimental design, and scientific conclusions are entirely the intellectual contribution of the human authors. All LLM-generated content underwent thorough human review and verification to ensure technical accuracy, scientific rigor, and alignment with our research objectives.

\section{Analysis for structured pruning Strategies}\label{app_comp}

This supplementary appendix is organized as follows. Section~\ref{app_comp} provides additional theoretical analysis for structured pruning, including the formal robustness statement, its proof, and the derivation of the pruning update rule. Section~\ref{app:exp_appendix} presents supplementary experimental details and extended empirical results, including task descriptions, additional commonsense reasoning results, pruning-strategy details, ablations, and full natural language understanding results.

In this section, we provide supplementary details and additional analysis complementing the methods discussion in the main paper. Appendix~\ref{app:analysis_two} presents the formal statement and proof of the robustness proposition for gradient-based structural pruning methods with respect to the overall loss. Furthermore, Appendix~\ref{appendix:solution_update} analyzes the minimizer of the main pruning objective and describes the procedure for pruning columns of a full weight matrix, as summarized in Algorithm~\ref{alg:gsp-obs}.

\subsection{Analysis for Gradient-based Pruning versus Activation-based Pruning}\label{app:analysis_two}

As discussed in the main paper, structured pruning strategies can be broadly categorized into two classes, both of which are widely adopted in foundation model compression~\citep{hubara2021accelerated, kurtic2025sparse, wu2024iterative, frantar2022optimal}. To better understand their implications, we provide a theoretical analysis examining how these strategies affect the overall loss. Since different approaches employ distinct criteria to measure precision, we first formalize the notion of perturbation error and analyze its influence on predictive performance. Let $\mW \in \mathbb{R}^{m\times n}$ denote the original weight matrix and $\widehat{\mW}$ its pruned counterpart. While our discussion primarily focuses on structured pruning, we note that our analysis, in principle, can be extended to non-structured settings.

It is important to highlight a key distinction between the two classes of methods for the sake of conceptual clarity. Although activation-based approaches can also apply a Taylor expansion and obtain the first-order gradient term, this gradient arises from the reconstruction objective rather than from the overall loss. In contrast, gradient-based pruning methods explicitly leverage the gradient of the overall loss, providing a more direct connection to the model’s predictive performance.
 
\begin{definition}[\textbf{$\varepsilon$-Perturbation Error}]\label{def_perturbation_error} We define the perturbation error under different pruning criteria as follows:
\begin{itemize}
    \item For \textbf{activation-based} pruning strategies, we say the pruned weight matrix \( \widehat{\mW} \) satisfies $\varepsilon$-perturbation error if: $\|\widehat{\mW} \mX - \mW \mX \|\leq \varepsilon$,
    where \( \mX \) is the input of the parameter layer.
    \item For \textbf{gradient-based} pruning strategies, we define $\varepsilon$-perturbation error as:
    $ |\mathcal{L}(\widehat{\mW}) - \mathcal{L}(\mW) | \leq \varepsilon$, where \( \mathcal{L} \) denotes the task-specific loss function.
\end{itemize}
\end{definition}

In Def.~\ref{def_perturbation_error}, the metrics of perturbation error for activation-based pruning and gradient-based pruning strategies derive from the activation-reconstruction objective and the gradient-based importance objective discussed in the main paper, respectively. Noticeably, even though we can set the same precision of the perturbation error for different pruning strategies (under Def.~\ref{def_perturbation_error}), we cannot know how the perturbation error of different pruning strategies contributes to the overall loss. Intuitively, gradient-based strategies emphasize preserving the global correlation between $\widehat{\mW}$ and $\mW$, which suggests greater robustness to weight perturbations for the overall loss. However, this intuition has not yet been formally established. In the following, we conduct an analysis on a single attention module to provide theoretical justification for this claim. It is an official statement of the robustness proposition introduced in the main paper.

\begin{proposition} (Official Statement) \label{proposition_perturbation_error}
    In a single attention module, if we assume each module of $(Q,K,V)$ satisfies perturbation error $\varepsilon$ in activation-based strategies, respectively, the overall loss would be linear w.r.t. the perturbation error up to the magnitude of each module. However, if they satisfy perturbation error $\varepsilon$ in gradient-based strategies, the overall loss would be linear in the perturbation error and independent of the magnitude of each module.
\end{proposition}

\textit{Proof:} \: Given an input $X \in \mathbb{R}^{n \times d_{\text{model}}}$, the query, key, and value module of a single attention module are obtained through three separate linear 
transformations:
\[
Q = X W_Q, \quad K = X W_K, \quad V = X W_V,
\]
where $W_Q, W_K, W_V \in \mathbb{R}^{d_{\text{model}} \times d}$ are 
trainable weight matrices, and $d$ is the dimensionality of a single 
attention head. Here, we assume these three modules have the same dimension. The attention output is then computed as
\[
Z = \text{softmax}\!\left(\frac{QK^\top}{\sqrt{d}}\right)V.
\]

The scaling factor $1/\sqrt{d}$ is introduced to prevent $QK^\top$ from growing too large in magnitude, 
which would otherwise make the softmax distribution extremely peaked and lead to unstable gradients. Given a weight vector $(x_1, x_2, \cdots, x_d)$, the softmax function will transform the $i$-th element in the vector as 
\[
\softmax(x_i) = \frac{\exp(x_i)}{\sum_j \exp(x_j)} ,
\]
which transforms a vector of real numbers into a probability distribution. 
In the attention mechanism, the softmax ensures that the attention weights assigned to all keys 
are non-negative and sum to one.

First, we will analyze activation-based pruning strategies. If we suppose $\|Q - \widehat{Q}\|_F\leq \varepsilon, \|K - \widehat{K}\|_F\leq \varepsilon, \|V - \widehat{V}\|_F \leq \varepsilon$, respectively, i.e., perturbation error in each module is bounded by $\varepsilon$ (See Def \ref{def_perturbation_error}). Then,
\[
\left\|Z - \widehat{Z}\right\|_F \leq \left\|A(V - \widehat{V})\right\|_F + \left\|(A - \widehat{A})\widehat{V}\right\|_F,
\]
where $A = \mathrm{softmax}\left(\frac{Q K^\top}{\sqrt{d}}\right)$ and $\widehat{A} = \mathrm{softmax}\left(\frac{\widehat{Q} \widehat{K}^\top}{\sqrt{d}}\right)$. The first term is at most $\varepsilon$ due to the fact that $\|A\| \leq 1$. The second term depends on the mismatch between $Q$ and $K$ after pruning:
\[
\left\|Q K^\top - \widehat{Q}\widehat{K}^\top\right\|_F \leq \left\|Q\right\| \cdot \left\|K - \widehat{K}\right\|_F + \|K\| \cdot \left\|Q - \widehat{Q}\right\|_F .
\]
This shows that the error in $A$ scales linearly with both $\varepsilon$ and the magnitude of $Q$ and $K$, leading to an overall bound:
\[
\left\|Z - \widehat{Z}\right\|_F \leq \left(1 + \frac{\|Q\| + \|K\|}{\sqrt{d}} \cdot \|\widehat{V}\| \right)\varepsilon
\]

In contrast, under the perturbation error of gradient-based tuning strategies, if we assume that $\mathcal{L}(Q,K,V)$ is the loss of a single attention module, we know that 
$$\left|\mathcal{L}(Q,K,V)-\mathcal{L}(\widehat{Q},\widehat{K},\widehat{V})\right| \leq 3 \varepsilon,$$
which is a direct consequence of the triangle inequality. This concludes the proof.

Next, we will analyze how pruning a single weight matrix $\mW$ affects the overall loss function $\mathcal{L}$ in the general cases.  
Assume that the loss function $\mathcal{L}$ is $C$-Lipschitz continuous (see \citep{federer2014geometric,latorre2020lipschitz} for formal definitions).  

For gradient-based pruning methods, if the pruning procedure introduces an $\varepsilon$-level perturbation error to the weights, the resulting loss change is at most $\varepsilon$, i.e., the approximation error in the loss is directly proportional to the perturbation error. This result is consistent with the conclusion we established on the toy model.

In contrast, for activation-based pruning methods, pruning a weight matrix with perturbation error $\varepsilon$ yields a change in the loss that is bounded by $C \varepsilon$, where $C$ is the Lipschitz constant of $\mathcal{L}$. Consequently, the sensitivity of the loss to perturbations induced by activation-based pruning can increase with the local sharpness of the loss landscape during fine-tuning, making its impact more difficult to control compared to gradient-based approaches.

Therefore, in the toy model, we can explicitly observe the impact of pruning multiple matrices under both gradient-based and activation-based strategies. The larger the matrix magnitude, the greater the error inflation in the overall loss function in activation-based methods. More generally, when considering a single weight matrix in any loss function, our analysis also highlights that activation-based methods are influenced by the Lipschitz constant, in contrast to gradient-based methods.

\subsection{Experiment for  Proposition 2}\label{App:pro2exp}

To empirically verify the robustness difference stated in Proposition~\ref{proposition_perturbation_error}, we conduct a controlled synthetic experiment using a self-attention model trained on a linear regression task. The goal is to examine how \emph{activation-based} and \emph{gradient-based} pruning strategies impact the change of loss under the same perturbation constraint~$\varepsilon$.

\paragraph{Data generation.}
We draw covariates $X\in\mathbb{R}^{n\times d}$ (rows are samples) i.i.d.\ from a zero-mean sub-Gaussian distribution (standard normal in our implementation), and generate responses
\[
y \;=\; X w^\star + \xi \;\in\; \mathbb{R}^{n\times 1},
\qquad
\xi \sim \mathcal{N}(0,\sigma^2 I_n),
\]
with $(n,d)=(2000,32)$. We use mean squared error (MSE) as the task loss:
\[
\mathcal{L}(\Theta)
\;=\;
\frac{1}{n}\,\big\| f_\Theta(X) - y \big\|_2^2.
\]

\paragraph{Model architecture.}
We use a single-layer self-attention model with a linear prediction head. Given an input matrix $X\in\mathbb{R}^{n\times d}$ with $n=2000$ and $d=32$, the model predicts scalar responses $\widehat{y}\in\mathbb{R}^{n\times 1}$ through three stages:

\begin{enumerate}
    \item \textbf{Input projection.}
    The raw features are first linearly projected to obtain $H\in\mathbb{R}^{n\times d}$:
    \[
    H = X W_{\text{in}} + \mathbf{1} b_{\text{in}}^\top,
    \]
    where $W_{\text{in}}\in\mathbb{R}^{d\times d}$ and $b_{\text{in}}\in\mathbb{R}^{d}$.
    
    \item \textbf{Single-head self-attention.}
    The attention block contains three trainable weight matrices:
    \[
    W_Q,\,W_K,\,W_V \in \mathbb{R}^{d\times d}.
    \]
    Queries, keys, and values are computed as
    \[
    Q = H W_Q,\qquad K = H W_K,\qquad V = H W_V.
    \]
    Attention weights are obtained via scaled dot-product:
    \[
    A = \softmax\!\left(\frac{Q K^\top}{\sqrt{d}}\right) \in \mathbb{R}^{n\times n},
    \]
    and the attention output is the weighted aggregation of values:
    \[
    Z = A V \in \mathbb{R}^{n\times d}.
    \]
    
    \item \textbf{Output projection.}
    The output head is a single linear transformation
    \[
    \widehat{y} = Z W_{\text{out}} + \mathbf{1} b_{\text{out}} \in \mathbb{R}^{n\times 1},
    \]
    with $W_{\text{out}}\in\mathbb{R}^{d\times 1}$ and $b_{\text{out}}\in\mathbb{R}$.
\end{enumerate}

The full parameter set is therefore
\[
\Theta = \{ W_{\text{in}}, b_{\text{in}}, W_Q, W_K, W_V, W_{\text{out}}, b_{\text{out}} \},
\]
which includes five linear weight matrices and two bias vectors. Then the model is trained using the mean-squared-error objective
\begin{align}\label{mse}
\mathcal{L}(\Theta) = \frac{1}{n}\big\|\widehat{y} - y\big\|_2^2.
\end{align}
optimized with Adam ($\text{lr}=10^{-3}$) for 2000 iterations until convergence.
The learned attention weights $\{W_Q,W_K,W_V,W_{\text{out}}\}$ are later used as the pruning targets in our robustness comparison experiments.

\paragraph{Activation-based pruning (SparseGPT-style).}
For each projection matrix $W$ with corresponding input activations $A_{\text{in}}$, we seek a sparse approximation $\widehat{W}$ that preserves the layer output within a small reconstruction error tolerance. Formally, we minimize the activation reconstruction discrepancy
\[
\mathrm{L}_{\text{act}}(A_{\text{in}},W,\widehat{W})
= \big\|A_{\text{in}}(W-\widehat{W})\big\|_F
\quad \text{s.t.} \quad
\mathrm{L}_{\text{act}}(A_{\text{in}},W,\widehat{W}) \le \varepsilon,
\]
where $\varepsilon$ controls the acceptable deviation in the forward activations.  
Practically, we solve a sequence of column-wise least-squares subproblems on $A_{\text{in}}$, greedily selecting the most significant columns of $W$ (in analogy to SparseGPT).  
We stop once the prescribed tolerance is reached.

\paragraph{Gradient-based pruning (LLM-Pruner-style).}
In contrast, gradient-based pruning leverages the first-order sensitivity of the loss function with respect to the model parameters.  
For each weight matrix $W\in\mathcal{W}$, we compute its gradient $G=\partial\mathcal{L}/\partial W$ and assign elementwise Taylor saliency scores
\[
s_{ij} = |W_{ij} G_{ij}|.
\]
Parameters with the smallest saliency scores are progressively pruned until the total parameter perturbation satisfies
\[
\mathrm{L}_{\text{w}}(W,\widehat{W})
= \big\|\mathcal{L}(W)-\mathcal{L}(\widehat{W})\big\|_F \le \varepsilon.
\]
This approach explicitly bounds the norm of the weight perturbation rather than the activation mismatch, ensuring that the induced change in loss remains linearly proportional to $\varepsilon$.

\paragraph{Results.}
Solving the prediction objective in (\ref{mse}) yields an estimated model with a baseline loss of $6.52\times 10^{-4}$. We then apply both pruning strategies under matched precision constraints.

For the \emph{gradient-based} method, enforcing the same perturbation budget produces pruned projection matrices that increase the loss only slightly, to $8.25\times 10^{-4}$. This corresponds to a negligible degradation, indicating that directly constraining the weight perturbation effectively preserves the model's predictive behavior.

In contrast, the \emph{activation-based} strategy yields a substantially larger post-pruning loss of $2.23\times 10^{-3}$, despite operating under an equivalent tolerance. This degradation---over three times larger than the gradient-based counterpart---highlights the instability of activation reconstruction as a pruning criterion: activation mismatch can propagate and be amplified through the network, making it markedly less robust under the same nominal precision level.

\subsection{Analysis for the Masking Pruning and Weight Update in the Main Pruning Objective}\label{appendix:solution_update}

In this part, we provide a detailed analysis of the main pruning objective:
\begin{align}
    \begin{split}
     \mathcal{M}_s, \mdelta = {\arg\min}_{\mathcal{M}_s,\mdelta} &  \left\langle \nabla_{\mW} \mathcal{L}, \mdelta \right\rangle 
+ \frac{1}{2} tr(\mdelta \widehat{\mH}\,\mdelta^\top )\\
     \textit{s.t.} \quad   &\mdelta_{:, \mathcal{M}_s} = - \mW_{:, \mathcal{M}_s}.
    \end{split}    
\end{align}
with optimal solutions for pruning selection  $ \mathcal{M}_s $ and weight update $ \mdelta $.

Here, for simplicity, we denote $\nabla_{\mW}\mathcal{L}(\mW)$ as $\nabla_{\mW}\mathcal{L}$. The corresponding Lagrange problem is 
\begin{align}\label{obj_1}
 \left\langle \nabla_{\mW} \mathcal{L}, \mdelta \right\rangle 
+ \frac{1}{2} tr(\mdelta \widehat{\mH} \mdelta^\top )
+  \langle \mLambda,   (\mdelta)_{:, \mathcal{M}_s} + \mW_{:, \mathcal{M}_s} \rangle,
\end{align}
where $\mLambda \in \mathbb{R}^{m\times n}$ is a Lagrange multiplier. Under first order condition of $\mdelta$, it implies 
\begin{align}\label{eq:first_order}
\nabla_{\mW} \mathcal{L} + \mdelta\widehat{\mH}  + \mLambda P_{\mathcal{M}_s} = 0.
\end{align}
where $P_{\mathcal{M}_s} \in \mathbb{R}^{n\times n}$ is a diagonal matrix whose $i$-th diagonal entry is $1$ if the $i$-th column is pruned and $0$ otherwise. Then we have 
\begin{align}
\mdelta = -  \left( \nabla_{\mW} \mathcal{L} + \mLambda P_{\mathcal{M}_s}\right) \widehat{\mH}^{-1} = - \nabla_{\mW} \mathcal{L} \: \widehat{\mH} ^{-1} -\mLambda  P_{\mathcal{M}_s} \:\widehat{\mH}^{-1}.
\end{align}

Then we can substitute the expression of $\mdelta$ back into the structured pruning constraint and obtain
\begin{align}\label{eq:lambda}
\mLambda = \left( \mW_{:,\mathcal{M}_s}-   (\nabla_{\mW} \mathcal{L} \: \widehat{\mH} ^{-1})_{:,\mathcal{M}_s}\right) \left( ( \widehat{\mH} ^{-1})_{\mathcal{M}_s,\mathcal{M}_s} \right)^{-1}.
\end{align}
Finally, putting the form of $\mLambda$ in (\ref{eq:lambda}) back into (\ref{eq:first_order}), we could get $\mdelta$ as 
\begin{align}\label{obj_update_app}
\begin{split}
\mdelta
&= - \nabla_{\mW} \mathcal{L} \: \widehat{\mH}^{-1} 
- \mW_{:, \mathcal{M}_s} 
\left( (\widehat{\mH}^{-1})_{\mathcal{M}_s, \mathcal{M}_s} \right)^{-1} 
(\widehat{\mH}^{-1})_{\mathcal{M}_s, :} \\
&\quad + (\nabla_{\mW} \mathcal{L}  \widehat{\mH}^{-1})_{:, \mathcal{M}_s} 
\left( (\widehat{\mH}^{-1})_{\mathcal{M}_s, \mathcal{M}_s} \right)^{-1} 
(\widehat{\mH}^{-1})_{\mathcal{M}_s, :}.
\end{split}
\end{align}

structured pruning methods \citep{liu2017learning,nova2023gradient} remove entire structured components of a network, facilitating efficient GPU speedups \cite{li2025sepprune}. Methods that utilize the gradient of the overall loss during training, which we refer to as gradient-based methods, can be more robust in controlling loss changes under weight perturbations induced by pruning. Gradients of the weights are computed during the standard optimization process, so they can be reused to estimate weight importance efficiently. Within the context of gradient-based pruning, we further explain the development of existing methods and clarify how our effort differs. Most works in the literature use an importance score to select the pruning structure \citep{molchanov2019importance, zhang2023adalora, shen2022structural,fang2023depgraph,molchanov2016pruning}. They provide refined pruning selection but do not further eliminate the influence of structured pruning. \citep{xia2022structured} combines distillation with pruning to improve performance and mitigate the impact of structured pruning, but it requires minimizing the KL-divergence of two distributions and does not admit a closed-form solution.

Inspired by Optimal Brain Surgeon, \citep{singh2020woodfisher,kurtic2022optimal, das2023beyond} propose a weight update after model pruning in the context of model compression to further eliminate the influence of pruning. Since their analysis is established for one-dimensional weight vectors, the pruning metric is hard to interpret. In contrast, we establish the analysis for the weight matrix and provide a grounded interpretation for the pruning selection and weight update, consistent with the main-paper discussion of gradient-based structured pruning.

\begin{algorithm}[t]
\caption{Gradient-Based Structured Pruning with Weight Update. We prune the layer matrix $\mW$ with column-wise sparsity $s$ given the gradient $\nabla_\mW \mathcal{L}$ and the Hessian matrix $\widehat{\mH} = (\nabla_{\mW} \mathcal{L})^T \nabla_{\mW} \mathcal{L}$}
\label{alg:gsp-obs}
\begin{algorithmic}[1]

\STATE \textbf{Step 1: Search pruning columns with sparsity $s$.} 
\[
\arg\min_{\mathcal{M}_s} \;
\mathrm{tr}\!\left(
(\mW - \nabla_\mW \mathcal{L}\:\widehat{\mH}^{-1})_{:,\mathcal{M}_s}
\left((\widehat{\mH}^{-1})_{\mathcal{M}_s,\mathcal{M}_s}\right)^{-1}
(\mW - \nabla_\mW \mathcal{L}\:\widehat{\mH}^{-1})_{:,\mathcal{M}_s}^\top
\right).
\]

\STATE \textbf{Step 2: Compute optimal update.} 
\STATE Given $\mathcal{M}_s$, compute update $\mdelta$:
\[
\mdelta = -\nabla_\mW \mathcal{L} \widehat{\mH}^{-1}
- (\mW-\nabla_\mW \mathcal{L}\widehat{\mH}^{-1})_{:,\mathcal{M}_s}
\big((\widehat{\mH}^{-1})_{\mathcal{M}_s,\mathcal{M}_s}\big)^{-1}
(\widehat{\mH}^{-1})_{\mathcal{M}_s,:}.
\]

\STATE \textbf{Step 3: Update model.} 
\STATE Set $\mW \gets \mW + \mdelta$.

\STATE \textbf{Step 4: Iterate or finalize.} 
\STATE If multi-round pruning, repeat Steps 1--3 until target sparsity/rank is reached. Otherwise, output $\mW$.

\end{algorithmic}
\end{algorithm}

\section{Experiment}\label{app:exp_appendix}

This appendix collects supplementary experimental details and extended empirical results. Appendix~\ref{app_nlgdes} describes the natural language generation tasks, training data, evaluation benchmarks, and compute setup used in the supervised fine-tuning experiments. Appendix~\ref{app:commonsense} reports additional commonsense reasoning results to test whether the gains of \textit{PrunedLoRA} generalize beyond mathematical reasoning and code generation. Appendix~\ref{app_prune_schedule} explains the pruning strategy and associated hyperparameter design, and Appendix~\ref{app_ablation} presents additional ablation studies.

\subsection{The Details about Natural Language Generation Task}\label{app_nlgdes}

\paragraph{Dialogue Generation Task}
We fine-tune large language models on a 52k subset of the WizardLM dataset~\cite{xu2024wizardlm} and evaluate it using the MT-Bench dataset~\cite{zheng2023judging}. GPT-4o is used to assess the quality of the model’s response and we report the first-turn score as the metric.

\paragraph{Math Task}
We fine-tune large language models on a 100k sample from the MetaMathQA dataset~\cite{yu2023metamath}. The model is then evaluated on the GSM8K test set~\cite{cobbe2021training}, and we report the accuracy as the metric.

\paragraph{Coding Task}
We fine-tune large language models on a 100k subset of the CodeFeedback dataset~\cite{zheng2024opencodeinterpreter} and test it on the HumanEval dataset~\cite{chen2021evaluating}, reporting the PASS@1 metric.

We fine-tune each task for three epochs, with a maximum of 5000 training steps. All experiments are conducted on NVIDIA H100 GPU cards.

\subsection{Commonsense Reasoning}\label{app:commonsense}

To further test whether the gains of \textit{PrunedLoRA} generalize beyond mathematical reasoning and code generation, we additionally evaluate it on eight commonsense reasoning benchmarks using the same LLaMA3-8B backbone. We report accuracy as the metric and compare LoRA with \textit{PrunedLoRA} under the same deployed target rank.

\begin{table*}[ht]
\centering
\small
\begin{threeparttable}
\resizebox{0.97\textwidth}{!}{%
\begin{tabular}{l|c|c|c|c|c|c|c|c|c}
\toprule
Method & BoolQ $\uparrow$ & PIQA $\uparrow$ & SIQA $\uparrow$ & HellaSwag $\uparrow$ & WinoGrande $\uparrow$ & ARC-e $\uparrow$ & ARC-c $\uparrow$ & OBQA $\uparrow$ & Avg. $\uparrow$ \\
\midrule
LoRA (rank = 8) & 69.24 & 83.38 & 78.80 & 88.32 & 84.13 & 83.62 & 71.22 & 79.32 & 79.75 \\
PrunedLoRA (init $r = 64 \rightarrow 8$) & 70.23 & 83.98 & 79.42 & 91.45 & 84.54 & 88.21 & 76.34 & 83.21 & 82.17 \\
\midrule
LoRA (rank = 64) & 71.23 & 85.88 & 80.37 & 92.34 & 85.60 & 89.43 & 80.43 & 84.59 & 83.73 \\
PrunedLoRA (init $r = 128 \rightarrow 64$) & 75.42 & 89.46 & 80.58 & 95.44 & 85.21 & 89.23 & 80.32 & 86.82 & 85.31 \\
\bottomrule
\end{tabular}%
}
\end{threeparttable}
\caption{Commonsense reasoning results on eight benchmarks using the LLaMA3-8B backbone. We report accuracy for each task and the average across tasks.}
\label{tab:commonsense_reasoning}
\end{table*}

Table~\ref{tab:commonsense_reasoning} shows that \textit{PrunedLoRA} consistently improves over standard LoRA in both rank regimes. In particular, the average score improves from 79.75 to 82.17 at target rank 8, and from 83.73 to 85.31 at target rank 64. This supports that the benefit of training in an over-parameterized adapter space and then compressing it into a compact final adapter extends to commonsense reasoning as well.

\subsection{Pruning Strategy}\label{app_prune_schedule}

\textbf{Dynamic Pruning.} Motivated by the main-paper observation that higher-rank LoRA adapters tend to achieve better empirical performance with smaller variance, we observe that higher-rank LoRA adapters ($\mA$ and $\mB$) provide a stronger starting point for subsequent structured pruning.
Based on this observation, we propose to prune adapters starting from higher-rank spaces. 
Specifically, we initialize adapters with rank $r \in \{128, 256, 512\}$ and progressively prune them down to rank $64$, corresponding to $50\%$, $75\%$, and $87.5\%$ sparsity, respectively. 
We also explore more aggressive settings (e.g., pruning from $r$ to $8$). 
Pruning is performed in a structured manner, controlled by two hyperparameters: the pruning interval $k_1$ and the number of columns removed per step $k_2$. 
For example, with $k_1 = 10$ and $k_2 = 2$, we prune two columns every ten training steps. 
Once the remaining columns reach the target rank budget (default: $64$), pruning is terminated.

\textbf{Adaptive Choice of Hyperparameter.} Importantly, as rank dynamically changes during training, the scaling factor $\alpha$ must remain stable. 
While vanilla LoRA typically sets $\alpha=16$, we find this choice suboptimal for higher-rank initializations. 
To address it, we perform a grid search over a large range and identify that $\alpha \in \{r/2, r, 2r\}$ can achieve the better performance, where $r$ is the current rank in LoRA. The hyperparameter $\alpha$ will be proportional to $r$ over the training process.

\subsection{Ablation Study}\label{app_ablation}

\begin{table}[ht]
\centering
\small
\begin{threeparttable}
\begin{tabular}{cccc}
\toprule
\textbf{Init Rank} & \boldmath{$\alpha$} & \textbf{GSM8K Acc.} & \textbf{Loss} \\
\midrule
128 & 64   &69.21 & 0.13 \\
128 & 128  & 71.11 & 0.14 \\
128 & 256  & 71.16 & 0.14 \\
\midrule
256 & 128  & 71.88 & 0.12 \\
256 & 256  & 72.21 & 0.11 \\
256 & 512  & 71.01 & 0.13 \\
\midrule
512 & 256  & 74.21 & 0.11 \\
512 & 512  & 74.21 & 0.10 \\
512 & 1024 & 73.99 & 0.12 \\
\bottomrule
\end{tabular}
\end{threeparttable}
\caption{Ablation study of \textit{PrunedLoRA} on GSM8K with different initial ranks and scaling factors $\alpha$ (rank/2, rank, 2$\times$rank). Each row reports Accuracy and the final training loss.}
\label{tab_ablation_alpha_init_r}
\end{table}

\textbf{Hyperparameter $\alpha$ and Initial Rank.} To better understand the sensitivity of \textit{PrunedLoRA} to the initial rank and the scaling factor $\alpha$, 
we conduct an ablation study on GSM8K with different settings of $\text{Init r} \in \{128, 256, 512\}$ 
and scaling factor $\alpha \in \{r/2, r, 2r\}$, where $r$ denotes the current rank. 
Table~\ref{tab_ablation_alpha_init_r} reports the results, with each row showing accuracy and loss. It shows that both the initialization rank and the scaling factor $\alpha$ play a critical role in the performance of \textit{PrunedLoRA}. For a fixed rank, setting $\alpha=r$ yields the best trade-off between accuracy and stability, while smaller values under-scale the updates and larger values bring little additional gain. Moreover, larger initialization ranks consistently improve results, with accuracy increasing from 72.21 at $r=128$ to 74.21 at $r=512$ when $\alpha=r$. These findings confirm that \textit{PrunedLoRA} benefits from high-rank initialization and that scaling $\alpha$ proportionally to the rank is the most effective choice.

\textbf{Comparison of Pruning Strategies under Different Initialization Ranks.}
Table~\ref{tab_sft_score_3_moarepruning} reports the performance of SparseGPT, LLM-Pruner, and PrunedLoRA with different initialization ranks ($r=128, 256, 512$). 
We observe that while all methods benefit from larger initial ranks, the gains are much more pronounced for \textit{PrunedLoRA}, which achieves the best performance at $r=512$. It further supports the effectiveness of gradient-based pruning over other structured pruning methods.

\begin{table}[ht]
\centering
\small
\begin{threeparttable}
\begin{tabular}{lccc}
\toprule
\textbf{Method} & \textbf{Init r} & \textbf{GSM8K} & \textbf{HumanEval} \\
\midrule
\multirow{3}{*}{\textit{SparseGPT} } 
  & 128 & 69.71$\pm$0.48 & 43.82$\pm$0.39 \\
  & 256 & 69.88$\pm$0.34 & 44.12$\pm$0.10 \\
  & 512 & 72.12$\pm$0.48 & 43.12$\pm$0.36 \\
\midrule
\multirow{3}{*}{\textit{LLM-Pruner} } 
  & 128 & 70.88$\pm$0.45 & 44.38$\pm$0.12 \\
  & 256 & 71.21$\pm$0.17 & 44.67$\pm$0.29 \\
  & 512 & \underline{74.19$\pm$0.23} & \underline{46.21$\pm$0.23} \\
\midrule
\multirow{3}{*}{\textit{PrunedLoRA}} 
  & 128 & 71.16$\pm$0.24 & 44.32$\pm$0.11 \\
  & 256 & 72.21$\pm$0.45  & 44.32$\pm$0.11 \\
  & 512 & \textbf{74.88$\pm$0.42} & \textbf{48.31$\pm$0.24} \\
\bottomrule
\end{tabular}
\end{threeparttable}
\caption{Comparison of \textit{SparseGPT}, \textit{LLM-Pruner}, and \textit{PrunedLoRA} under different initial ranks on GSM8K and HumanEval benchmarks using Llama-3-8B-Base. \textbf{Bold} indicates the best result, \underline{underline} represents the second-best one.}
\label{tab_sft_score_3_moarepruning}
\end{table}

\textbf{Pruning Schedule $K_1$ and $K_2$.} 
We further investigate the impact of the pruning schedule on the performance of \textit{PrunedLoRA}. 
Specifically, we vary the pruning interval $K_1 \in \{5, 10\}$, which controls how frequently pruning is applied, 
and the number of columns pruned at each step $K_2 \in \{2, 4\}$. 
Table~\ref{tab:ablation_pruning_schedule} summarizes the results on GSM8K. 
We find that less frequent pruning with a smaller number of pruning indices at each pruning step (e.g., $K_1=10$, $K_2=2$) leads to stable performance, 
while larger $K_2$ values slightly hurt accuracy. It suggests that gradual pruning with moderate intervals achieves better performance.

\begin{table}[ht]
\centering
\small
\begin{threeparttable}
\begin{tabular}{ccc}
\toprule
\textbf{$K_1$} & \textbf{$K_2$} & \textbf{GSM8K} \\
\midrule
5  & 2 & 69.39 \\
5  & 4 & 70.23 \\
10 & 2 & 71.16 \\
10 & 4 & 71.12 \\
\bottomrule
\end{tabular}
\end{threeparttable}
\caption{\textit{PrunedLoRA} on GSM8K with different pruning schedules. 
$K_1$ is the pruning interval (steps between pruning), and $K_2$ is the number of pruning indices at each step.}
\label{tab:ablation_pruning_schedule}
\end{table}

\begin{figure}[ht]
    \centering
    \subfloat[Init $r=64$]{%
        \includegraphics[width=0.45\linewidth]{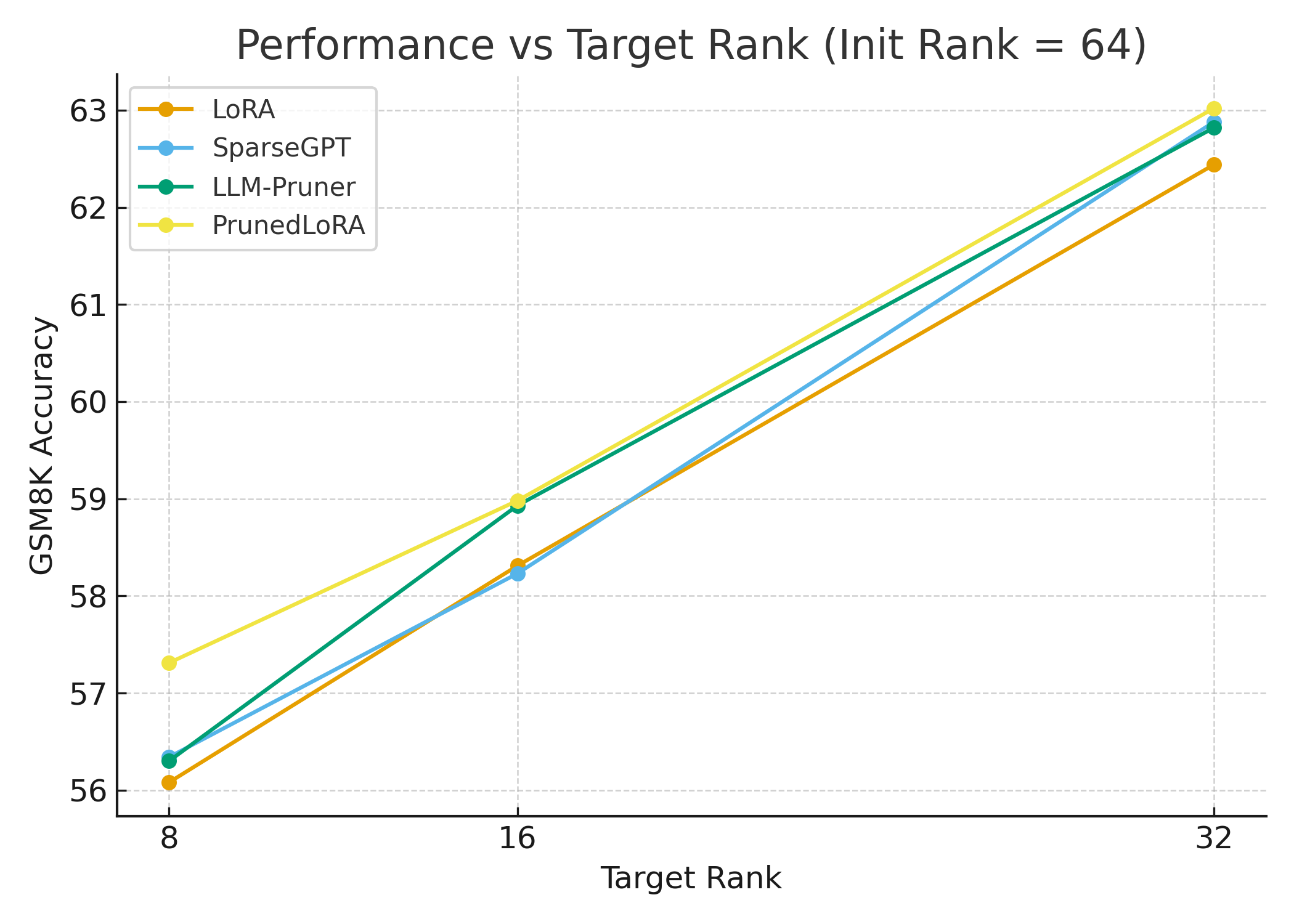}}
    \hfill
    \subfloat[Init $r=128$]{%
        \includegraphics[width=0.45\linewidth]{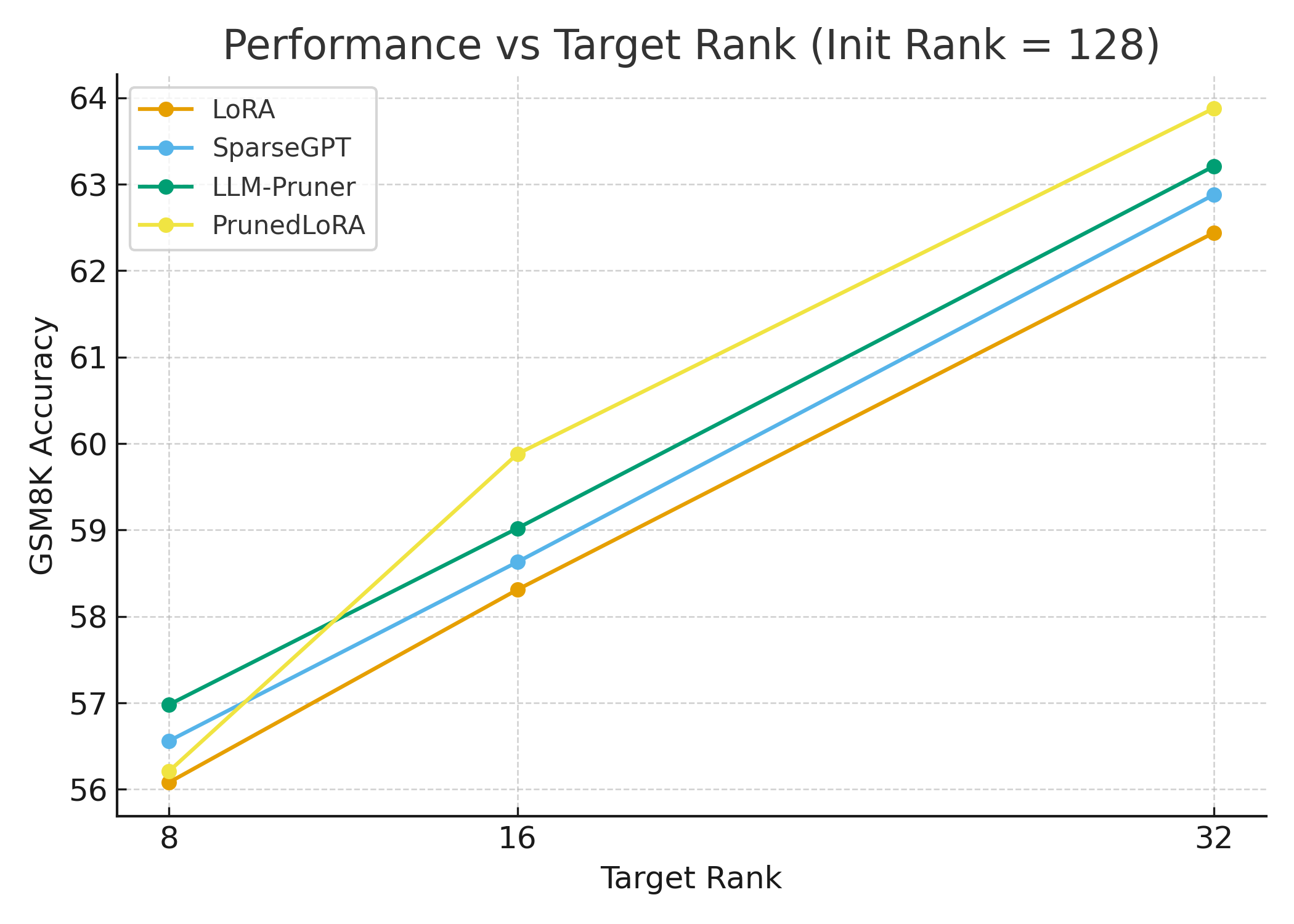}}
    \\
    \subfloat[Init $r=256$]{%
        \includegraphics[width=0.45\linewidth]{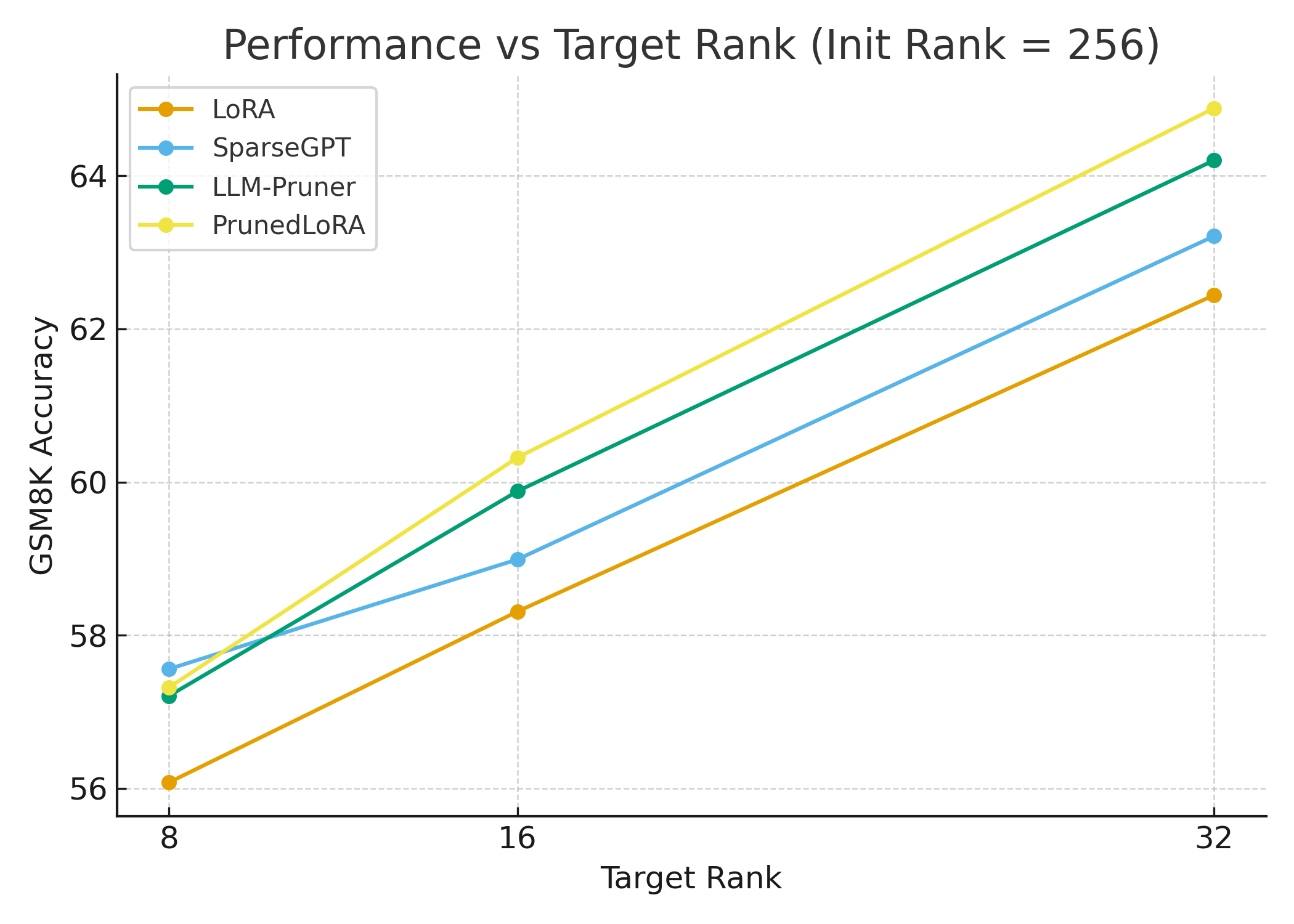}}
    \hfill
    \subfloat[Init $r=512$]{%
        \includegraphics[width=0.45\linewidth]{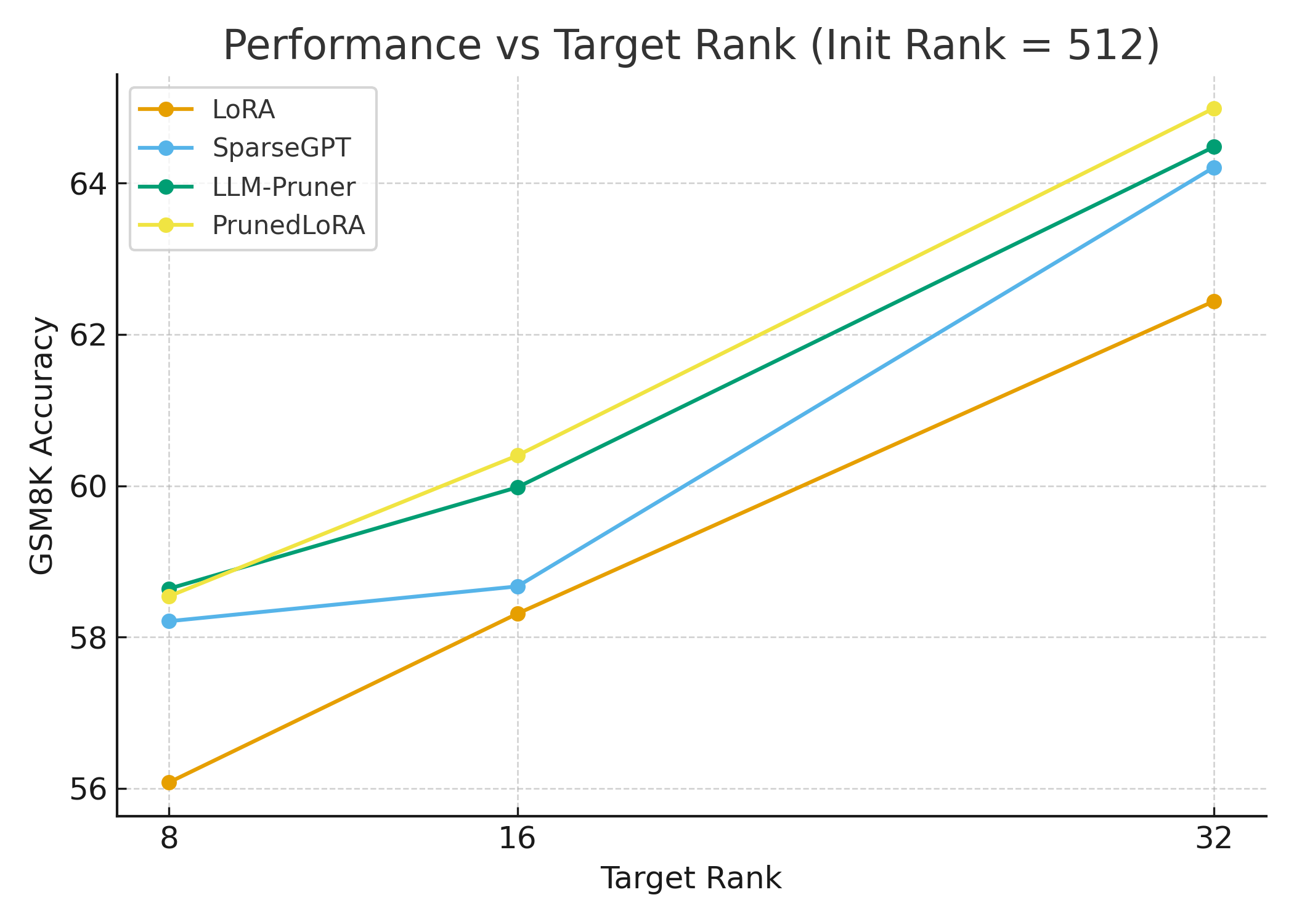}}
    \caption{GSM8K accuracy of different pruning methods (SparseGPT, LLM-Pruner, and PrunedLoRA) under various initialization ranks $r \in {64,128,256,512}$ and target ranks ${8,16,32}$. Each subfigure reports performance when starting from a specific initialization rank.}
    \label{fig:low_rank_pruning}
\end{figure}

\textbf{Pruning for Different Low-Rank Targets.} We further investigate the effect of initialization rank and pruning budget on downstream performance. Figure~\ref{fig:low_rank_pruning} presents results where LoRA adapters are initialized with $r={512, 256, 128, 64}$ and pruned to smaller target budgets ($r= 32, 16, 8$). Across all settings, \textit{PrunedLoRA} consistently outperforms classical one-shot pruning approaches such as SparseGPT and LLM-Pruner, and maintains accuracy close to or above the unpruned LoRA baseline. The performance gap becomes more pronounced when the pruning ratio is high (e.g., pruning LoRA from init $r=128$ to target rank 8), highlighting that gradient-informed structured pruning is more robust under extreme compression. These results confirm that \textit{PrunedLoRA} provides both stability and generalization, making it preferable when adapting to stringent memory and efficiency constraints.

\section{Additional Natural Language Understanding Results}
\label{app:glue_results}

We provide the full GLUE benchmark results referenced in the main paper. Consistent with the main generation experiments, \textit{PrunedLoRA} improves over LoRA-style baselines under moderate training budgets and further narrows the gap to full fine-tuning when larger initialization ranks are affordable.

\begin{table*}[t]
\centering
\small
\begin{threeparttable}
\setlength{\tabcolsep}{4pt}
\begin{tabular}{l|c|c|c|c|c|c}
\toprule
Method & MNLI $\uparrow$ & SST2 $\uparrow$ & CoLA $\uparrow$ & QNLI $\uparrow$ & MRPC $\uparrow$ & Average $\uparrow$ \\
\midrule
Full FT & \underline{86.33$\pm$0.06} & \underline{94.75$\pm$0.21} & \textbf{80.70$\pm$0.24} & 93.19$\pm$0.22 & \textbf{84.56$\pm$0.73} & \textbf{87.91} \\
LoRA    & 85.30$\pm$0.04 & 94.04$\pm$0.11 & 69.35$\pm$0.05 & 92.96$\pm$0.09 & 68.38$\pm$0.01 & 82.08 \\
\midrule
DoRA    & 85.67$\pm$0.09 & 94.04$\pm$0.53 & 72.04$\pm$0.94 & 93.04$\pm$0.06 & 68.08$\pm$0.51 & 82.57 \\
AdaLoRA & 85.45$\pm$0.11 & 93.69$\pm$0.20 & 69.16$\pm$0.24 & 91.66$\pm$0.05 & 68.14$\pm$0.28 & 81.62 \\
\midrule
SparseGPT & 85.21$\pm$0.23 & 93.33$\pm$0.19 & 68.16$\pm$0.34 & \underline{94.33$\pm$0.15} & 73.32$\pm$0.34 & 82.07 \\
LLM-Pruner & 84.76$\pm$0.12 & 93.12$\pm$0.30 & 65.21$\pm$0.25 & 93.39$\pm$0.33 & 76.43$\pm$0.31 & 82.18 \\
PrunedLoRA (init r = 128) & 85.21$\pm$0.32 & 93.21$\pm$0.29 & 73.43$\pm$0.23 & 93.34$\pm$0.12 & 74.21$\pm$0.18 & 83.48 \\
PrunedLoRA (init r = 256) & 86.21$\pm$0.09 & 94.21$\pm$0.31 & 74.43$\pm$0.32 & \textbf{94.55$\pm$0.05} & 78.21$\pm$0.28 & 85.12 \\
PrunedLoRA (init r = 512) & \textbf{86.67$\pm$0.12} & \textbf{95.22$\pm$0.34} & \underline{78.43$\pm$0.45} & 93.45$\pm$0.25 & \underline{84.19$\pm$0.34} & \underline{87.19} \\
\bottomrule
\end{tabular}
\end{threeparttable}
\caption{GLUE benchmark results with different adaptation methods. Best results are in \textbf{bold}, second-best are \underline{underlined}. ($\uparrow$: higher values indicate better performance).}
\label{tab:glue_results}
\end{table*}

\end{document}